\documentclass[sigconf, 10pt, nonacm]{acmart}
\AtBeginDocument{%
  \fancypagestyle{firstpagestyle}{%
    \fancyhf{}
    \fancyhead[C]{\Small
      PREPRINT: Accepted to The Tenth ACM/IEEE Symposium on Edge Computing (SEC'25).\quad
      Final version: \url{https://doi.org/10.1145/3769102.3770622}}
    
  }%
}

\usepackage{adjustbox}
\usepackage{amsmath,amsfonts}
\usepackage{graphicx}
\usepackage{textcomp}
\usepackage{xcolor}
\usepackage{booktabs}
\usepackage{multirow}
\usepackage{algorithm}
\usepackage{algpseudocode}
\usepackage{subcaption}
\usepackage{makecell}
\usepackage{url}
\usepackage{pifont}
\newcommand{\cmark}{\ding{51}}
\newcommand{\xmark}{\ding{55}}

\begin{document}

\title{\textsc{PlatformX}: An End-to-End Transferable Platform for Energy-Efficient Neural Architecture Search}

\author{Xiaolong Tu}
\email{xtu1@student.gsu.edu}
\affiliation{%
  \institution{Georgia State University}
  \city{Atlanta}
  \state{GA}
  \country{USA}}

\author{Dawei Chen}
\email{dawei.chen1@toyota.com}
\affiliation{%
 \institution{Toyota InfoTech Labs}
  \city{Mountain View}
  \state{CA}
  \country{USA}}

\author{Kyungtae Han}
\email{kt.han@toyota.com}
\affiliation{%
  \institution{Toyota InfoTech Labs}
  \city{Mountain View}
  \state{CA}
  \country{USA}}

\author{Onur Altintas}
\email{onur.altintas@toyota.com}
\affiliation{%
  \institution{Toyota InfoTech Labs}
  \city{Mountain View}
  \state{CA}
  \country{USA}}
  
\author{Haoxin Wang}
\email{haoxinwang@gsu.edu}
\affiliation{%
  \institution{Georgia State University}
  \city{Atlanta}
  \state{GA}
  \country{USA}}

\renewcommand{\shortauthors}{Xiaolong Tu et al.}

\begin{abstract}
Hardware-Aware Neural Architecture Search (HW-NAS) has emerged as a powerful tool for designing efficient deep neural networks (DNNs) tailored to edge devices. However, existing methods remain largely impractical for real-world deployment due to their high time cost, extensive manual profiling, and poor scalability across diverse hardware platforms with complex, device-specific energy behavior.

In this paper, we present \textsc{PlatformX}, a fully automated and transferable HW-NAS framework designed to overcome these limitations. \textsc{PlatformX} integrates four key components: (i) an energy-driven search space that expands conventional NAS design by incorporating energy-critical configurations, enabling exploration of high-efficiency architectures; (ii) a transferable kernel-level energy predictor across devices and incrementally refined with minimal on-device samples; (iii) a Pareto-based multi-objective search algorithm that balances energy and accuracy to identify optimal trade-offs; and (iv) a high-resolution runtime energy profiling system that automates on-device power measurement using external monitors without human intervention. 
% Together, these components form a closed-loop pipeline for scalable and accurate energy-aware model search.
We evaluate \textsc{PlatformX} across multiple mobile platforms, showing that it significantly reduces search overhead while preserving accuracy and energy fidelity. It identifies models with up to 0.94 accuracy or as little as 0.16 mJ per inference, both outperforming MobileNet-V2 in accuracy and efficiency. 
% These results demonstrate that \textsc{PlatformX} enables scalable, energy-aware HW-NAS for sustainable AI deployment on heterogeneous edge devices.
% You can find our code, demo, and tutorials at \url{https://github.com/amai-gsu/PlatformX}.
Code and tutorials are available at \href{https://github.com/amai-gsu/PlatformX}{\color{magenta}\texttt{github.com/amai-gsu/PlatformX}}.
\end{abstract}

% \begin{CCSXML}
% <ccs2012>
%    <concept>
%        <concept_id>10010520.10010553</concept_id>
%        <concept_desc>Computer systems organization~Embedded and cyber-physical systems</concept_desc>
%        <concept_significance>500</concept_significance>
%        </concept>
%    <concept>
%        <concept_id>10010147.10010257</concept_id>
%        <concept_desc>Computing methodologies~Machine learning</concept_desc>
%        <concept_significance>500</concept_significance>
%        </concept>
%  </ccs2012>
% \end{CCSXML}

% \ccsdesc[500]{Computer systems organization~Embedded and cyber-physical systems}
% \ccsdesc[500]{Computing methodologies~Machine learning}

\keywords {Sustainable Edge AI, Energy-Efficient DNNs, Automated NAS Platform.}

\maketitle

\section{Introduction}

Hardware-Aware Neural Architecture Search (HW-NAS) has emerged as a powerful paradigm for automatically designing deep neural networks (DNNs) optimized for specific hardware platforms \cite{wu2019fbnet, dong2020nasbench201, li2021hwnasbench}. However, applying NAS to build energy-efficient models remains fundamentally challenging for two reasons. First, NAS requires evaluating millions of candidate architectures, making the process computationally expensive and labor-intensive \cite{tan2019efficientnet, gu2020brpnas}
Second, accurate energy-aware evaluation demands fine-grained power profiling \cite{tu2023unveiling, wang2020energy}, which is inherently device-dependent and often relies on external power monitors. This complexity is particularly problematic for edge devices, where inference completes in milliseconds. 
% especially critical for edge devices where model inference completes in mere milliseconds. 
This reliance adds substantial setup complexity and hinders the scalability of NAS pipelines across heterogeneous edge platforms \cite{yang2022knas, canziani2019teadnn}.

\textit{Large-Scale Model Search Remains Costly and Manual.}
To identify high-performing models, NAS must often explore extensive search spaces that capture a broad spectrum of architectural possibilities. While such expansiveness is essential for discovering competitive designs, it imposes steep computational costs and engineering overhead. 
Early NAS methods based on reinforcement learning requires supercomputer-scale budgets. The original NASNet \cite{pham2018efficient,zoph2018learning}, for instance, search consumes roughly 32,000 to 75,000 GPU-hours (equivalent to running 450 GPUs for three to seven days) merely to search on CIFAR-10 before even attempting transfer to larger datasets like ImageNet.
Evolutionary search algorithms push the resource demands even higher. AmoebaNet-A \cite{real2019regularized} reports 75,600 GPU-hours (3,150 GPU-days) to achieve comparable accuracy. 
In response to these expensive computational demands, weight-sharing techniques such as ENAS \cite{chu2021tunas, mellor2021naswot} has been proposed, which can reduce the search cost to fewer than 16 GPU-hours on a single GTX 1080 Ti. However, these methods require carefully tuned one-shot supernets and compromise training fidelity, leading to suboptimal transferability across tasks and datasets. 
Moreover, simply reducing the raw GPU-hours does not eliminate the need for manual intervention. Instead, they shift it toward the design of the supernet, the tuning of search algorithms, and the adaptation to specific hardware platforms.

\textit{Platform-aware NAS methods introduce a new axis of complexity.}
MnasNet \cite{tan2019mnasnet}, for instance, integrates real latency measurements into its reward function, requiring over 40,000 GPU-hours and millions of device-side inference traces. This is motivated by the observation that traditional proxies like FLOPs and parameters often fail to accurately reflect real-world latency behavior. 
ProxylessNAS~\cite{cai2019proxylessnas, feng2024litepred} mitigates this by training latency predictors to guide the search process, reducing the compute cost to approximately 200 GPU-hours. However, these predictors must be retrained and revalidated whenever the target hardware or firmware changes, limiting their robustness in dynamic or heterogeneous deployment settings.
More recent frameworks, such as Once-for-All \cite{cai2020onceforall}, Pi-NAS \cite{liu2021pilot}, and CE-NAS \cite{zhao2024cenas}, further amortise search cost across tasks, but they still depend on high-quality device profiling to avoid accuracy and energy regressions.  

\textit{Accurate Energy Profiling Requires Intrusive Instrumentation.}
Energy profiling \cite{tu2023deepen2023, tu2025greenauto, tu2025aienergy} is essential for designing energy-efficient models, as developers must understand real power consumption before optimization.
On mobile and embedded platforms, fine-grained power measurements typically require external instrumentation, such as Monsoon Power Monitor, which replaces battery leads on smartphones. These methods often involve hardware modifications, physical access to power rails, and custom scripts for streaming voltage and current samples~\cite{boubouh2023powerprofiler}.
To avoid invasive setups, many studies rely instead on easily obtainable proxy metrics \cite{mallik2023epam}, such as FLOPs~\cite{howard2017mobilenets}, latency~\cite{cai2018proxylessnas}, or hardware performance counters~\cite{yang2018netadapt}. However, these proxies often show weak correlation with actual energy consumption, especially on modern platforms that feature Dynamic Voltage and Frequency Scaling (DVFS), shared memory bandwidth, and runtime kernel fusion \cite{lin2020mcunet, canziani2019teadnn}. 
Consequently, they fail to capture critical system-level effects like thermal throttling, cache contention, and kernel fusion. 
Frameworks that include power measurement in the loop often suffer from coarse temporal resolution and limited automation. For instance, KNAS records board-level average power once per inference batch and requires manual triggering by an engineer for each run. These limitations make existing power profiling workflows unsuitable for scalable or real-time NAS scenarios, where evaluating millions of model candidates demands high-throughput, fully automated pipelines.

To this end, we propose \textsc{PlatformX}, a fully automated, end-to-end HW-NAS framework designed to optimize energy-efficient models for diverse edge platforms. \textsc{PlatformX} integrates model search, deployment, runtime measurement, and adaptive optimization into a unified and scalable pipeline. It consists of four core components:
\begin{itemize}
    \item \textbf{Energy Efficiency-driven Search Space.} We extend conventional NAS search spaces, which often prioritize accuracy alone, by incorporating energy-critical architectural configurations, such as kernel sizes, channel dimensions, and stride. This expanded design space promotes the exploration of architectures that that balance accuracy with improved energy efficiency.
    
    \item \textbf{Transferable Kernel-level Energy Prediction.} Instead of retraining a new energy predictor for each target device, \textsc{PlatformX} introduces a transferable kernel-level energy prediction framework that generalizes across devices. The predictor is initialized from prior measurements and fine-tuned using a small number of device-specific samples, substantially reducing profiling overhead when adapting to new hardware. To the best of our knowledge, \textsc{PlatformX} represents the first transferable kernel-level energy prediction design for edge devices.
    % It initializes from existing measurements and is fine-tuned using a small number of on-device samples. This significantly reduces the profiling and training overhead when adapting to new hardware platforms.
    
    \item \textbf{Multi-objective Optimization for Model Search.} We formulate HW-NAS as a multi-objective optimization problem jointly optimizing accuracy and energy consumption. A gradient-based Pareto search algorithm guides the exploration toward the Pareto frontier, guaranteeing that the selected architectures represent true optimal trade-offs. The search operates in an iterative, measurement-driven loop, continuously refined with real-device feedback so that the resulting architectures capture genuine hardware behavior rather than relying on proxy metrics.
    
    \item \textbf{Automated Model Runtime Performance Profiling.} \textsc{PlatformX} integrates a high-resolution, fully automated energy profiling pipeline that measures inference-time power using external monitoring tools. The profiling system operates without human intervention and is tightly coupled with the search and prediction modules, enabling real-time feedback and iterative model refinement. While most existing HW-NAS approaches target datacenter GPUs or server environments, \textsc{PlatformX} is explicitly designed and validated for resource-constrained edge devices (e.g., smartphones, Jetson), where millisecond-level profiling accuracy and energy efficiency are critical.
\end{itemize}

Unlike prior HW-NAS systems that rely on coarse analytical proxies or labor-intensive profiling, 
\textsc{PlatformX} closes the loop with millisecond-resolution, on-device energy feedback, enabling robust and deployable model selection. 
As shown in Table~\ref{tab:search_cost}, \textsc{PlatformX} achieves the lowest total search cost while offering the highest profiling granularity among representative HW-NAS systems. 
Specifically, \textsc{PlatformX} completes the full search in only 7 GPU-days, more than 400$\times$ faster than NASNet-A (3,150 GPU-days) and nearly 10$\times$ faster than TEA-DNN (9.7 days), 
despite operating over a much larger search space (\(9.6 \times 10^{5}\) vs. \(2 \times 10^{4}\) models). 
Moreover, unlike prior methods that measure only coarse model-level or proxy metrics, 
\textsc{PlatformX} performs kernel-level, millisecond-scale energy profiling with full automation and real on-device measurements---capabilities not supported by NASNet-A, TEA-DNN, BRP-NAS, or MnasNet. 
This combination of fine-grained measurement fidelity, end-to-end automation, and real-device validation enables \textsc{PlatformX} to significantly reduce search time while maintaining hardware-accurate energy estimation, demonstrating both scalability and practical deployability across edge platforms.

We evaluate \textsc{PlatformX} across a range of mobile and embedded devices, including both CPU- and GPU-based platforms. Results demonstrate that \textsc{PlatformX} consistently achieves high energy prediction accuracy with minimal calibration, identifies models with optimal energy--accuracy trade-offs, and reduces profiling overhead by an order of magnitude compared with state-of-the-art alternatives. 
In summary, \textsc{PlatformX} bridges the gap between scalable neural architecture search and practical energy-aware optimization. By unifying adaptive prediction, fully automated profiling, and Pareto-based search, it provides an effective and deployable solution for designing high-performance, energy-efficient models for edge deployment.

\begin{table*}[t]
  \centering
  \caption{Comparison of evaluation costs and profiling capabilities across representative HW-NAS systems. Only \textbf{PlatformX} supports fine-grained energy measurement, full automation, and real on-device profiling, significantly reducing search cost.}
  \label{tab:search_cost}
  \resizebox{\textwidth}{!}{
  \begin{tabular}{lllllccc}
    \toprule
    \textbf{Method} & \textbf{Search Space Size} & \textbf{Objectives} & \textbf{Wall-clock Time} & \textbf{\# Models Profiled} & \textbf{Energy Granularity} & \textbf{Auto Search} & \textbf{On-device} \\
    \midrule
    \textbf{PlatformX (Ours)} & $9.6 \times10^5$ & Accuracy, Latency, Energy & 7 GPU-days & 170 & Kernel-level (ms) & \cmark~Full loop & \cmark \\
    NASNet-A \cite{pham2018efficient} & $2 \times10^4$ & Accuracy & 3,150 GPU-days & 20,000 & \xmark & \xmark & \xmark \\
    TEA-DNN \cite{cai2019tea} & $5.6 \times 10^{14}$ & Accuracy, Energy & 9.7 days & 400 & Model-level (s) & \xmark & \xmark \\
    BRP-NAS \cite{dudziak2020brp}& $10^7$ & Accuracy, Latency & up to 60 days & — & \xmark & \xmark & \xmark \\
    MnasNet \cite{tan2019mnasnet}& $8 \times10^3$ & Accuracy, Latency & 4.5 days & 8,000 & \xmark & \xmark & \cmark \\
    \bottomrule
  \end{tabular}
  }
\end{table*}

\section{Background and Motivation}
\label{sec:background}

\subsection{Bottlenecks to Energy-Efficient HW-NAS}
\label{subsec:bottlenecks}
HW-NAS offers a principled route to energy-efficient models on resource-constrained edge devices. Its practical deployment, however, is curtailed by three interrelated bottlenecks.

\textbf{Bottleneck 1 - Escalating evaluation cost.} The dominant obstacle to deployable energy-efficient HW-NAS is the sheer cost of evaluation. Each surviving candidate architecture must (i) be partially or fully trained to obtain a statistically meaningful accuracy estimate, (ii) undergo repeated inference runs to measure latency under realistic batch sizes and DVFS settings, and (iii) be profiled at millisecond resolution to capture power and energy behavior on edge devices. These stages incur substantial GPU hours, prolonged device occupation, and significant engineering overhead.
Critically, evaluation time has not scaled down in proportion to the rapid expansion of NAS search spaces. 
The field has evolved from NASNet-A’s \(2\times10^{4}\) candidates to TEA-DNN’s \(5.6\times10^{14}\) alternatives—an increase of ten orders of magnitude—yet per-model evaluation remains prohibitively expensive. 
TEA-DNN still allocates roughly \(25\,\mathrm{min}\) of training and \(10\,\mathrm{min}\) of high-frequency power tracing \emph{per candidate}; evaluating just \(400\) filtered models therefore consumes about \(9.7\) GPU-days on a single V100 GPU. BRP-NAS demonstrates an even steeper trend: its breadth-first progressive pruning evaluates \(10^{7}\) architectures, monopolizing a 112-GPU cluster for 60 consecutive days before the Pareto set stabilizes.
These data reveal a super-linear scaling law: once per-architecture evaluation exceeds a few minutes, even moderate increases in search-space cardinality inflate total wall-clock time by orders of magnitude, rendering exhaustive exploration infeasible. 
Consequently, reducing the evaluation budget without sacrificing the discovery of promising architectures remains the key bottleneck for scalable and sustainable HW-NAS.

\textbf{Bottleneck 2 - Inefficient multi-objective exploration.}
Once the per-candidate cost is contained, search strategy itself becomes the limitations for energy-efficient HW-NAS.
An effective algorithm should identify promising architectures under multiple conflicting objectives (accuracy, latency, energy) before expensive evaluation begins, otherwise the search still degenerates into a prohibitively large test-and-discard loop.
Current practice remains largely accuracy-centric. The original MnasNet controller drew a reinforcement-learning policy over 8,000 candidates, but 98\% were thrown away once the on-device latency constraint was enforced, dissipating nearly all of the optimization effort on infeasible designs. ProxylessNAS replaced explicit measurements with a differentiable latency estimator and trimmed the raw compute budget to 200 GPU hours; yet when energy was added as a third objective, hundreds of latency-compliant architectures were still discarded because they violated the energy bound.  
These case studies illustrate a fundamental inefficiency: sequentially optimizing one objective and filtering on the others yields a combinatorial explosion in wasted trials as soon as a new metric is introduced.  Without an efficient multi-objective mechanism that steers the search directly toward the direction, the number of rejected architectures and the associated computational cost escalate super-linearly with each additional design constraint.

\textbf{Bottleneck 3 - Scarcity of cross-device, high-resolution energy profiling.}
Inference on edge devices is bursty and short-lived \cite{wang2020user, wang2022leaf+, wang2017v}; individual kernels often finish within milliseconds.  
Capturing energy at a matching granularity is therefore essential, because coarse or perturbed traces obscure true power behaviour and mislead hardware-aware search.  
Unfortunately, system-level counters such as Android \texttt{BatteryStats} and \texttt{nvidia-smi} update at sub-hertz rates and miss these events entirely.  
Although instrumenting privileged kernel hooks can raise the sampling rate, the additional interrupts disturb DVFS governors, pre-empt user tasks, and contaminate the very measurements they aim to record.
Consequently, most HW-NAS studies resort to external instrumentation.  
TEA-DNN couples the target phone to a 1 kHz USB power meter, yet publishes only an average joule per model number, forfeiting kernel-level insight. MCUNet adds a shunt resistor and INA226 sensor to every micro-controller board; soldering, gain calibration, and temperature stabilisation require 1 to 2 hours per device.  
External monitors such as the Monsoon Power Monitor are a another common way to obtain high-granularity energy data without adding runtime overhead to the mobile device. However, setting up the monitor requires disassembling the phone to remove the battery, based on our experience, preparation and voltage calibration take about two hours per device and void the hardware warranty.
The need for labour-intensive, platform-specific measurement pipelines severely limits the volume and diversity of energy data available for HW-NAS.  
Without an automated, high-resolution, and broadly transferable profiling method, it is impossible to scale energy-aware search across the heterogeneous of modern edge hardware.

\subsection{Design Opportunities}
\label{subsec:design_opportunities}
Addressing the three bottlenecks demands an end-to-end strategy that unifies efficient screening, multi-objective exploration, and autonomous, high-fidelity measurement.
First, an energy-aware search space, coupled with transferable energy predictors, must prune the evaluation pipeline and guide the search toward low-power solutions. With accurate kernel-level estimates available up front, only a small, high-quality subset of architectures proceeds to full training and on-device profiling.
Second, the search algorithm must evaluate accuracy, latency, and energy from outset. By expressing HW-NAS as a continuous multi-objective optimisation problem that quantifies trade-offs in real time, the search concentrates on architectures that promise the best overall balance, thereby eliminating superfluous model evaluations and saving compute resources.
Finally, reliable, high-granularity energy profiling without human intervention is essential. An automated pipeline that streams millisecond-resolution traces delivers trustworthy energy labels across heterogeneous hardware while avoiding the hours of manual setup that currently hinder large-scale experimentation.
By integrating these elements, a HW-NAS framework can cut wall-clock search time by an order of magnitude, scale to trillions of candidate networks, and remain portable across the rapidly evolving landscape of edge devices.

\section{System Design}

\begin{figure*}[t]
  \centering
  {\includegraphics[width=0.98\linewidth]{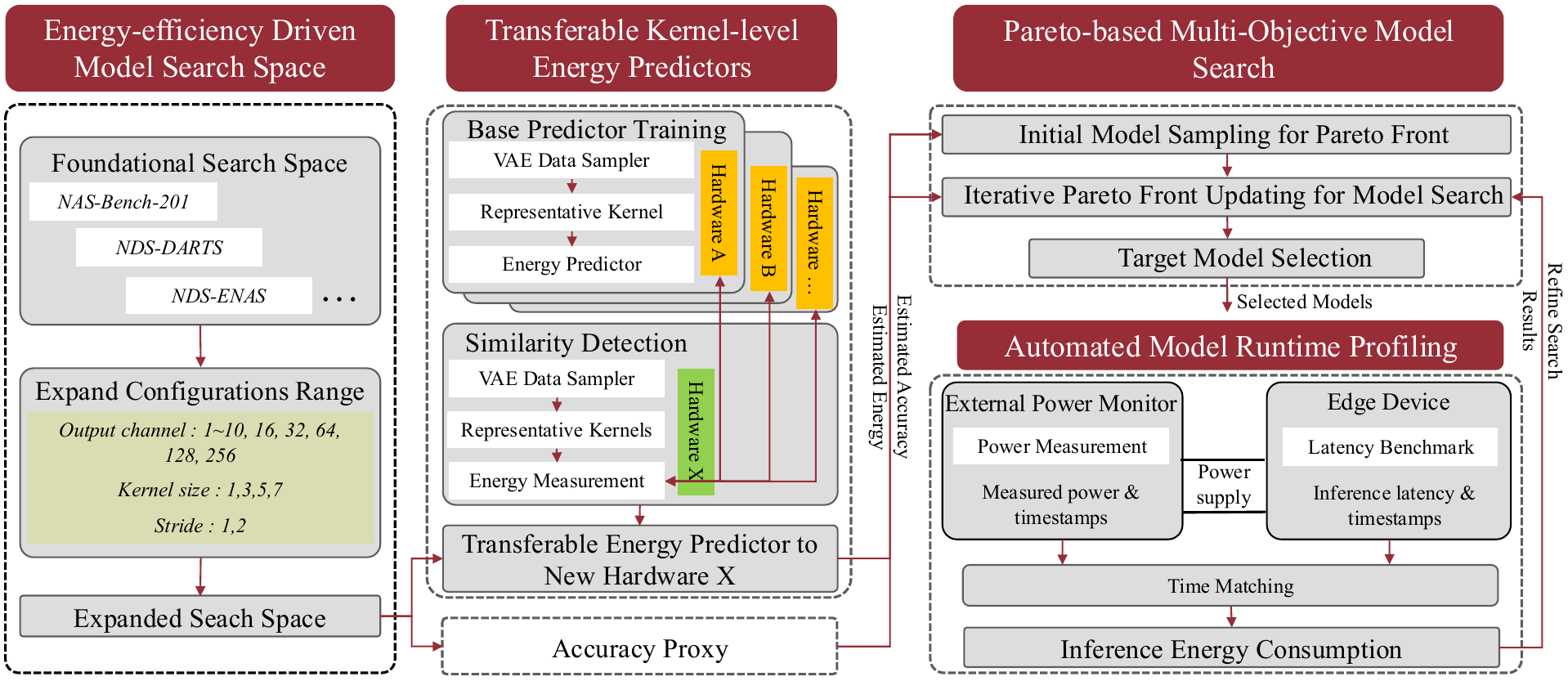}}
  \caption{System overview of \textsc{PlatformX}, an automated platform for energy-efficient NAS. It integrates energy-efficiency driven search space generation, transferable energy prediction, Pareto-based model search, automated on-device energy profiling.}
  \label{fig:system_pipeline}
  % \vspace{-0.15in}
\end{figure*}

The observations in Section~\ref{sec:background} motivate the design of \textsc{PlatformX}, a fully automated system designed to enable scalable and energy-aware neural architecture search across diverse edge platforms. As illustrated in Fig.~\ref{fig:system_pipeline}, the system integrates four core components: energy efficient search space generation, kernel-level energy prediction, Pareto-based model search, and automated runtime measurement into a streamlined. The entire process is self-updating and requires no manual intervention. Given a foundational NAS search space, \textsc{PlatformX} automatically outputs energy-efficient models optimized for the target hardware.

The system begins by constructing an \textit{energy-efficiency driven search space} (\S\ref{sec:search_space}) that encodes architectural factors known to influence energy consumption. In contrast to conventional NAS spaces focused solely on accuracy or latency, \textsc{PlatformX} explicitly includes energy-relevant attributes such as kernel size, input/output channel dimensions, and stride. These design parameters are chosen based on their strong empirical correlation with energy usage, as determined through prior profiling across multiple platforms.

Each candidate model in the search space is then evaluated using a \textit{Transferable Kernel-level energy predictors} (\S\ref{ssc:predictor}), which estimates energy consumption without requiring full-model deployment. This predictor is trained using individual kernel configurations, represented by their operator type along with associated structural parameters such as convolution kernel size, input channels, and output channels. By focusing on the kernel level, the predictor captures energy behavior that is more consistent across different models and hardware platforms compared to full-network profiling. This modular design enables efficient and accurate energy estimation while maintaining generalization across diverse architectures. As new measurements are collected from real-device executions, the predictor is incrementally updated to better reflect hardware-specific energy characteristics.

To efficiently explore the search space, \textsc{PlatformX} employs a \textit{Pareto-based multi-objective model search} (\S\ref{ssc:Pareto}). Rather than treating energy as a hard constraint or combining it with accuracy via scalarization, the system identifies models that lie near the Pareto front. This approach prioritizes architectures that offer the best trade-offs between accuracy and energy consumption. In each iteration, a small number of promising models are selected for actual deployment, guided by the gradients that inform beneficial search directions.
Once models are selected, they are deployed to the target edge platform for direct evaluation using the \textit{automated model runtime profiling} (\S\ref{ssc:measure}) subsystem. This module includes a power monitor connected to the device's power rails and a benchmark execution script that performs inference on the selected model. During inference, high-frequency energy samples are recorded and aligned with operator execution traces to compute total energy consumption with fine temporal resolution. The results of this measurement are used to validate the predictor's estimates and to log true performance metrics for each candidate.
Finally, the system uses the real measurements to update the predictor and search process. Kernel configurations extracted from the evaluated models are re-profiled individually if their energy prediction errors exceed a threshold. These measurements are appended to the predictor's training data, and the predictor is retrained to reflect the updated hardware behavior. The new predictions are then propagated across the entire search space, ensuring that subsequent candidate selection is based on the most accurate available estimates.

Through these components, \textsc{PlatformX} forms a tightly integrated, self-correcting pipeline for hardware-aware NAS. It drastically reduces the need for manual profiling, avoids exhaustive search, and adapts to different edge devices with minimal per-device calibration. This design enables fast, accurate, and energy-efficient model discovery suitable for real-world deployment across heterogeneous platforms.

\subsection{Energy Efficiency-driven Search Space}
\label{sec:search_space}

\textsc{PlatformX} expands conventional NAS search spaces to prioritize energy efficiency, addressing the limitations of accuracy-centric designs. As a foundation, we build upon NAS-Bench-201 \cite{dong2020bench}, a widely used cell-based architecture benchmark composed of three stacked cell groups connected via residual blocks. Each stack contains five repeated cells (\(N=5\)), with each cell represented as a densely-connected directed acyclic graph (DAG) of four nodes. The original search space includes five operation candidates: zeroize, skip connection, 1x1 convolution, 3x3 convolution, and 3x3 average pooling, yielding a total of \(5^6 = 15{,}625\) unique architectures.

However, NAS-Bench-201 fixes several architectural parameters, limiting its utility for energy-aware design. For instance, the input and output channel sizes (\(C_{in}, C_{out}\)), stride (\(S\)), and kernel size (\(KS\)) are statically set—e.g., \(C_{out}\) is fixed at 16, 32, and 64 across the three stacks, and \(KS\) is limited to 1 or 3. Prior work \cite{tu2023unveiling} has shown that such architectural configurations have a strong influence on energy consumption, making rigid parameterization suboptimal for practical deployment on energy-constrained edge devices.

To address this, \textsc{PlatformX} systematically expands the search space by relaxing key architectural constraints, as summarized in Table~\ref{tb:nastable}. This includes a broader range of output channels, additional kernel sizes, and variable stride values. Although this expansion theoretically yields over one million candidate models, many configurations are invalid due to shape mismatches, unsupported operations, or compilation failures. To ensure executable candidates, \textsc{PlatformX} performs rigorous validation, including shape compatibility checks and TFLite-based compilation filtering, resulting in a curated and executable pool of 959,417 valid models.
Exhaustive search over this large space is computationally impractical. Profiling every valid model on a single device would take more than eight years of continuous testing, as reported in prior work. To make this process tractable, \textsc{PlatformX} incorporates two fast and scalable estimation strategies. First, it uses our pre-trained kernel-level energy predictor, which is introduced in the following subsection (\ref{ssc:predictor}), and estimates energy consumption based on operator configurations without requiring full-model deployment. Second, it employs a zero-cost proxy for accuracy estimation, following the approach proposed in NASWOT \cite{mellor2021neural}. By combining these two estimators, the system reduces the evaluation time to approximately 85 milliseconds per model. In contrast, conventional approaches that involve full training and power measurement typically require over 1.2 hours per model. This improvement makes it feasible to explore the expanded search space efficiently and at scale.

\begin{table}[t]
\centering
\caption{Convolutional layer configuration: original vs. expanded}
\begin{adjustbox}{max width=\columnwidth}
\footnotesize
% \resizebox{0.46\textwidth}{!}{
\begin{tabular}{l|l|l}
\toprule
Configuration & Original Space & Expanded Space \\
\hline    
Initial Input Channels & 16 & 16 \\\hline
Output Channels & 16, 32, 64 & 1-10, 16, 32, 64, 128, 256 \\\hline
Kernel Size & 1, 3 & 1, 3, 5, 7 \\\hline
Stride & 1 & 1, 2 \\
\bottomrule
\end{tabular}
% }
\label{tb:nastable}
\end{adjustbox}
\end{table}

\subsection{Transferable Kernel-Level Energy Predictors}
\label{ssc:predictor}

Efficient energy estimation is critical in the early stages of the NAS pipeline. Reliable predictions enable \textsc{PlatformX} to identify promising model candidates, filter out low-potential designs, and significantly reduce overall search cost. However, constructing predictors that generalize across heterogeneous hardware platforms remains a key challenge.

To address this, \textsc{PlatformX} adopts a strategy inspired by LitePred, beginning with synthetic data generation using a variational autoencoder (VAE). For each kernel, the VAE produces 1,000 configurations. These configurations are instantiated as micro-benchmarks and executed on both the CPU and GPU of the target device. Each run captures the tensor shape (\(H \times W\)), input channels (\textit{Cin}), output channels (\textit{Cout}), kernel size (\textit{KS}), stride (\(s\)), and the energy consumption measured in joules using an external power monitor. The resulting traces are used to train a compact multilayer perceptron (MLP), which serves as the base energy predictor.

When \textsc{PlatformX} is deployed on a new hardware platform, the base predictor may no longer yield accurate estimates due to variations in DVFS, cache hierarchy, and power management policies. Instead of collecting millions of new samples for each target device, the system performs lightweight few-shot calibration. Specifically, it profiles only 50 to 100 representative kernels on the new hardware, compares the observed energy values with predictions, and selects the base predictor with the smallest Kullback-Leibler divergence. This selected model is then fine-tuned using the small set of newly acquired samples.
Calibration effort is prioritised where it matters most. Empirical profiling shows that the convolutional patterns \texttt{conv+bn+relu} and \texttt{dwconv+bn+relu} account for 93.97\% and 87.74\% of total energy on mobile CPUs and GPUs, respectively, and that kernel size dominates energy cost owing to its quadratic relationship with compute complexity. \textsc{PlatformX} therefore allocates a larger share of its calibration budget to these operators and to configurations with diverse kernel sizes, while spending fewer samples on parameters (e.g., channel widths, stride) whose effects scale nearly linearly.

This targeted calibration approach enables the construction of accurate, device-specific energy predictors using only a fraction of the data required by traditional profiling methods. It provides a scalable and adaptive foundation for energy estimation in cross-platform NAS.

\subsection{Pareto-based Multi-Objective Model Search}
\label{ssc:Pareto}

With energy and accuracy estimates available, \textsc{PlatformX} formulates model exploration as a multi-objective optimization problem \cite{gunantara2018review}. The goal is to discover models that minimize energy consumption while maintaining high accuracy:

\begin{equation}
    \min F(x) = (f_{1}(x), f_{2}(x), ..., f_{m}(x)), \quad x \in \Omega,
\end{equation}

\noindent where \(x\) is a DNN model from the search space \(\Omega\), and each objective \(f_i(x)\) represents a quantity to minimize—here, estimated energy \(C_p\) and NASWOT score \(N_s\) (as a proxy for accuracy).
Model \(x_1\) is said to \emph{Pareto dominate} \(x_2\) (denoted \(x_1 \prec x_2\)) if:
\begin{equation}
    \left\{
    \begin{aligned}
    &\forall i, \ f_i(x_1) \leq f_i(x_2); \\
    &\exists j, \ f_j(x_1) < f_j(x_2).
    \end{aligned}
    \right.
\end{equation}

The Pareto front \cite{kang2024survey} \(\mathcal{P}\) contains all such non-dominated models. To efficiently traverse this front, \textsc{PlatformX} applies a gradient-descent-based sampling strategy. Rather than evaluating all models, the system selectively refines regions of the search space where trade-offs between accuracy and energy are promising.
In each iteration, a small batch of models on or near the current Pareto frontier is selected and deployed to the target device. Real energy measurements are collected and used to validate predictions, refine the Pareto front, and update the predictor if necessary. This targeted exploration maximizes discovery efficiency, ensuring that only the most promising architectures are evaluated and improved upon.

% \subsection{Pareto-based Multi-Objective Search Strategy}
% \label{sec:carbon_search}

\textsc{PlatformX} incorporates real-world sustainability constraints into model search by minimizing energy consumption alongside maintaining high accuracy. 
% The total carbon emissions \(C\) are estimated as:
% \begin{equation}
%     C = E \times CI,
%     \label{eq:carbon_emissions}
% \end{equation}
% where \(E\) is the energy consumption of a model and \(CI\) is the carbon intensity (e.g., kgCO\(_2\)/kWh) associated with the regional power grid.
To align the optimization objectives, we normalize energy and accuracy estimates prior to search. Energy consumption \(C_p\) is normalized using Min-Max scaling, and the NASWOT-based accuracy score \(N_s\) is log-normalized:
\begin{equation}
\begin{aligned}
C_{p_{\text{norm}}} &= \frac{C_p - \min(C_p)}{\max(C_p) - \min(C_p)}, \\
N_{s_{\text{norm}}} &= \log(N_s).
\end{aligned}
\label{eq:normalization}
\end{equation}

\textbf{Initial Model Sampling for Pareto Front.} PlatformX begins by sampling \(k\) diverse candidate models based on the distribution of normalized \(C_p\) and \(N_s\). These models are trained to determine real accuracy and measured on-device for actual energy usage, forming an initial Pareto front.

\textbf{Iterative Pareto Front Updating for Model Search.} To refine the Pareto frontier iteratively, \textsc{PlatformX} uses a multi-objective gradient descent (MGD) strategy. Let \(g_i(x)\) denote the gradient of the \(i\)-th objective for model \(x\), the optimal gradient direction \(g^*(x)\) is calculated as:
\begin{equation}
g^*(x) \propto \sum_{i=1}^{m} \lambda_i^*(x) g_i(x),
\end{equation}
where \(\{\lambda_i^*(x)\}_{i=1}^{m}\) solve:
\begin{equation}
\begin{aligned}
\min_{\{\lambda_i\}} \left\| \sum_{i=1}^{m} \lambda_i g_i(x) \right\| \quad 
\text{s.t.} \quad \sum_{i=1}^{m} \lambda_i = 1, \ \lambda_i \geq 0.
\end{aligned}
\end{equation}
New candidates are selected based on the alignment between their gradient \(g(x)\) and the optimal direction \(g^*(x)\), using the inner product as a similarity measure. In each iteration, the top-\(m\) aligned models are evaluated on the target hardware, and the Pareto front is updated. This process repeats until a model satisfies the predefined constraints.
To enable scenario-specific tuning, we introduce weighting coefficients \(ws_i^*(x)\) for each objective, which scale \(\lambda_i^*\) to bias the search. For example, higher \(ws_\text{energy}\) emphasizes low-power models.
Algorithm~\ref{algorithm:Pareto_search} summarizes the Pareto optimization process.

\begin{algorithm}[t]
  \caption{Pareto Optimization with Gradient-Guided Sampling}
  \label{algorithm:Pareto_search}
  \begin{algorithmic}[1]
    \Statex \textbf{Input:} Normalized energy $C_p(x)$, accuracy $N_s(x)$
    \State Sample $k$ initial models from search space $\mathcal{X}$
    \State Train and measure real energy/accuracy for these $k$ models
    \State Build initial Pareto front $\mathcal{P}_0$
    \State Compute initial gradient direction $g^{\!*}(x)_0$
    \While{no $x\in\mathcal{P}$ satisfies the constraints}
      \State Select $m$ models most aligned with $g^{\!*}(x)$
      \State Measure new models and update $\mathcal{P}$
      \State Recompute $g^{\!*}(x)$
    \EndWhile
    \Statex \textbf{Output:} Final Pareto front $\mathcal{P}^{\!*}$
  \end{algorithmic}
\end{algorithm}

% \subsection{Carbon-Efficient Model Selection}
% \label{sec:carbon_model_selection}

\textbf{Target Model Selection.} Once the Pareto front is stabilized, \textsc{PlatformX} identifies the best trade-off model using a weighted gradient descent approach. The best model \(x^*\) is chosen to have the smallest weighted gradient norm, allowing tunable preference toward accuracy or energy objectives:
\begin{equation}
\text{Select } x^* = \arg\min_{x \in \mathcal{P}} \left\| \sum_{i} wd_i \cdot g_i(x) \right\|.
\end{equation}
Algorithm~\ref{algorithm:BestModelSelection} outlines the model selection process.
\begin{algorithm}[t]
\caption{Gradient-Based Selection of Optimal Model}
\label{algorithm:BestModelSelection}
\begin{algorithmic}[1]
    \State \textbf{Input:} Pareto front $\mathcal{P}$, weights $wd_i$
    \State Initialize $x^* = \emptyset$, $\min\_grad = \infty$
    \For{each $x \in \mathcal{P}$}
        \State Compute gradients $g_i(x)$ and weighted $g_i^w = wd_i \cdot g_i(x)$
        \State Calculate magnitude $\|g^w(x)\|$
        \If{$\|g^w(x)\| < \min\_grad$}
            \State Update $x^* = x$
        \EndIf
    \EndFor
    \State \textbf{Output:} Best model $x^*$
\end{algorithmic}
\end{algorithm}
This dynamic adjustment makes \textsc{PlatformX} capable of optimizing models not only for performance but also for real-world sustainability targets across heterogeneous hardware.

\subsection{Automated Model Runtime Profiling}
\label{ssc:measure}

As discussed in \S\ref{ssc:Pareto}, iterative refinement of the Pareto front in \textsc{PlatformX} relies on accurate, on-device energy measurements and model accuracy evaluations. To address this, \textsc{PlatformX} incorporates an external high-resolution power monitor, enabling automated profiling and reducing reliance on manual instrumentation, which is often time-consuming and error-prone.

During the evaluation phase, model inference runs on the edge device while the external power monitor continuously samples power data. Each inference invocation records its start and end timestamps, \(T_s\) and \(T_e\), on the edge device. Meanwhile, the power monitor samples electrical current and voltage at high frequency, each tagged with a timestamp \(T_m\).
However, aligning \(T_s\) and \(T_e\) with the timestamps of the power monitor (\(T_m\)) poses synchronization challenges. Although both the edge device and the monitor are synchronized to the host machine, clock drift and latency variations across models make static sampling windows unreliable.
To ensure accurate measurement, \textsc{PlatformX} adopts an event-based synchronization strategy. Inference start and stop events trigger and terminate the power data capture process, ensuring that power samples comprehensively cover the model inference duration. From the sampled data, \textsc{PlatformX} identifies the corresponding power samples between \(T_s\) and \(T_e\), computes the average current and voltage, and calculates the average power and total energy using:
\[
P = I \cdot V, \quad E = P \cdot (T_e - T_s).
\]

For model accuracy evaluation, \textsc{PlatformX} performs training and validation on the host server using task-specific datasets, such as CIFAR-10 \cite{krizhevsky2009learning}. This approach ensures that energy measurements and accuracy assessments are both robust and scalable across large candidate sets.

\section{System Implementation}
\label{sec:system}

\textsc{PlatformX} is a fully automated hardware-software co-design platform built from four physical elements: a GPU server, an Android device, a high-frequency power monitor, and a dedicated Wi-Fi router.  
The components work in concert to generate candidate networks, deploy them, capture fine-grained power traces, and drive multi-objective optimization.

\begin{figure}[t]
  \centering
  {\includegraphics[width=\linewidth]{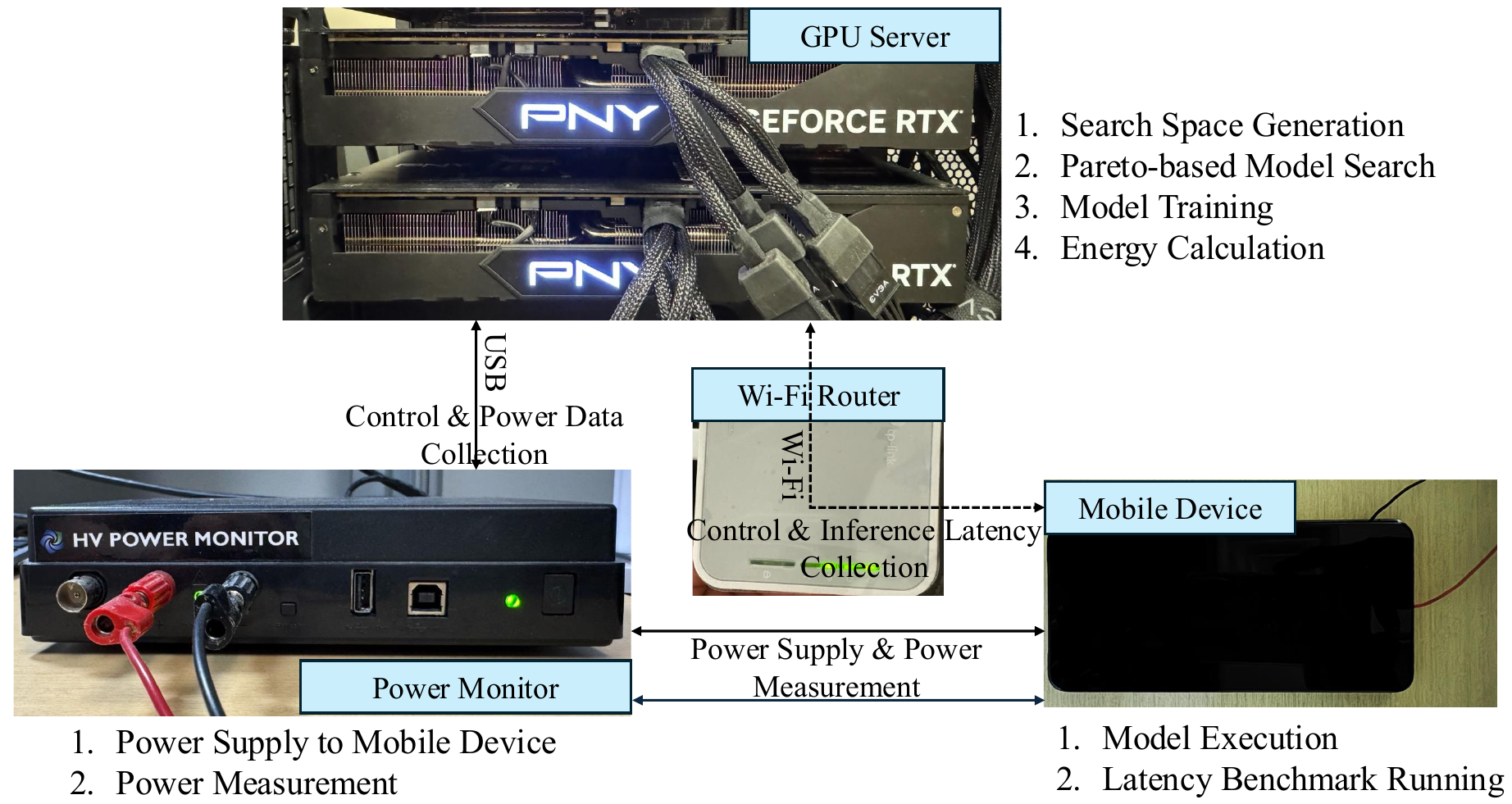}}
  \caption{Platform setup and the functional allocation of \textsc{Platform X}.  }
  % \Description{system implementation}
  \label{fig:implementation}
\end{figure}

\subsection{\textsc{PlatformX} Platform Setup}

The Platform setup of \textsc{Platform X} is illustrated in Fig. \ref{fig:implementation}. There are four hardware used in the platform, GPU server, mobile device, Power monitor, and Wi-Fi router. 

\textbf{GPU host server.}  
The server orchestrates the entire search loop.  
It generates architectures, runs kernel-level energy and accuracy predictors, trains selected models, and converts them to TensorFlow-Lite format.  
All logs and measurement results are archived locally to support iterative optimisation.  
An NVIDIA RTX 4090 GPU provides training throughput and establishes ground-truth accuracy.  
The server also synchronises benchmark execution and data collection by communicating with both the phone and the power meter.

\textbf{Mobile device.}  
Deployments target a Android device equipped with an ARM CPU and a Mali GPU.  
To obtain uncontaminated power readings the internal battery is removed and the phone is powered directly from the external monitor, avoiding USB-charging artefacts.  
Models are pushed over Wi-Fi and executed with the TensorFlow-Lite benchmark tool \cite{tflitebenchmarktool}; ADB carries control commands and log files.  
For reproducibility we minimise screen brightness, disable adaptive settings, and stop all background services and radios.

\textbf{Power monitor.}  
A Monsoon Power Monitor \cite{Monsoon} samples voltage and current at 5{kHz}, capturing the sub-millisecond energy fluctuations typical of edge inference.  
Compared with software counters, the Monsoon delivers hardware-level accuracy for power and cumulative joule counts.

\textbf{Wi-Fi router.}  
A dedicated router isolates traffic between server and handset, eliminating congestion that could otherwise skew deployment latency or trigger timing.

\subsection{Timing Synchronization Between Edge Device and Meter.}  

\begin{figure}[t]
  \centering
  {\includegraphics[width=\linewidth]{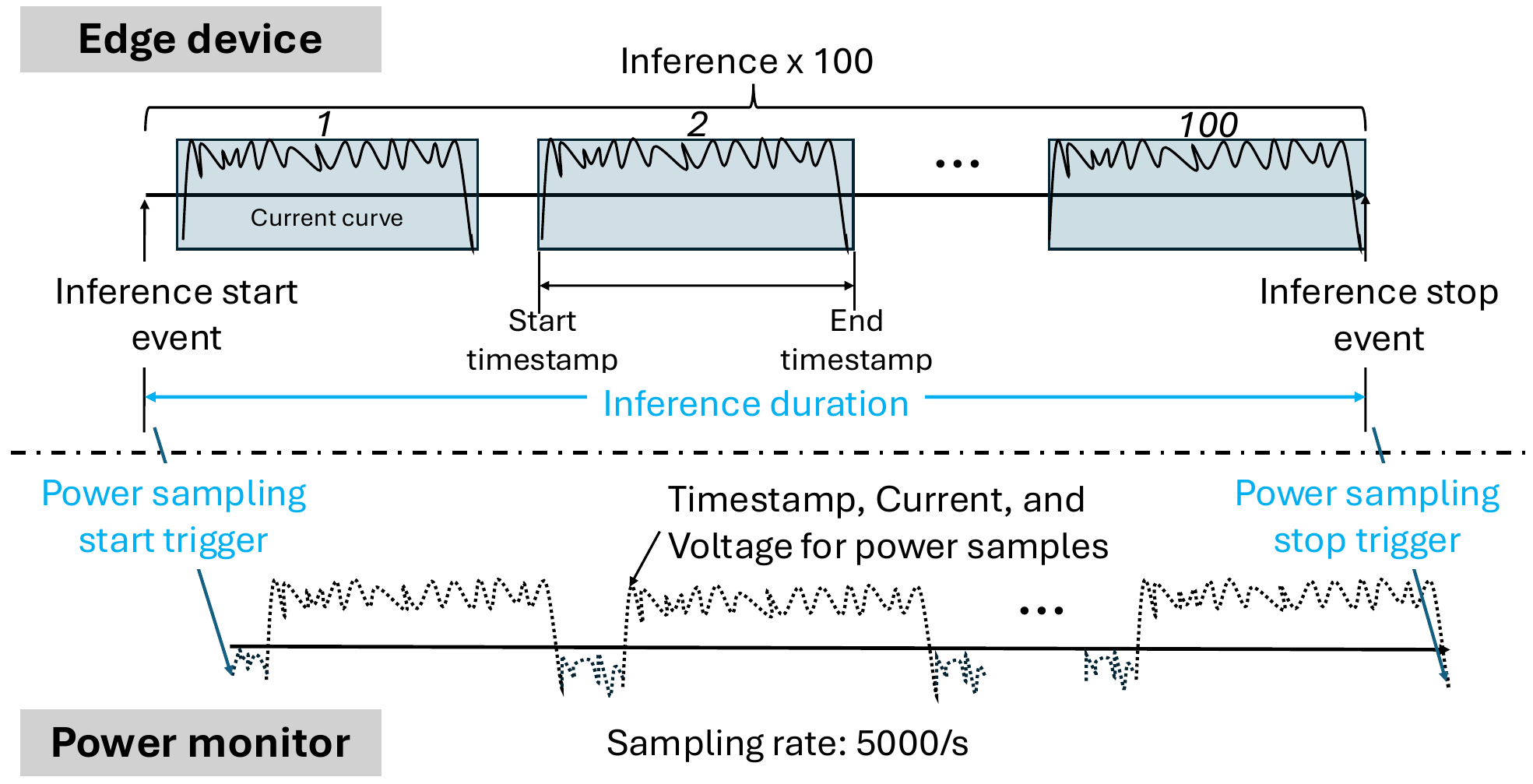}}
  \caption{Timing sync between edge device and external power monitor.}
  % \Description{system pipeline}
  \label{fig:time sync}
\end{figure}

Model inference occurs on the phone, whereas power is sampled by the Monsoon; aligning the two timelines is essential.  
As Fig.~\ref{fig:time sync} shows, the handset records a start time~\(T_s\) and end time~\(T_e\) for each inference.  
Concurrently, the monitor emits timestamped samples \(\langle T_m,\,I,\,V\rangle\) at a fixed rate.  
Even with a shared clock, drift can accumulate during long runs, and inference latency varies from model to model, so a fixed sampling window is unsafe.

\textsc{PlatformX} therefore drives the meter with two explicit triggers: an \emph{inference-start} event that begins power capture and an \emph{inference-stop} event that ends it.  
Using \(T_s\) as an anchor, the system selects the first power sample whose timestamp exceeds \(T_s\) and then all subsequent samples up to \(T_e\), guaranteeing full coverage of the execution interval.  
Average current and voltage are computed from this slice; multiplying by latency yields average power and total energy.

This automated workflow eliminates manual instrumentation, provides high-resolution ground truth, and feeds the adaptive search loop without human intervention.

\section{Evaluation}

\subsection{Experiment Setup}
We conduct experiments on three widely used NAS benchmarks: \textsc{NAS-Bench-201}, \textsc{NDS-ENAS}, and \textsc{NDS-DARTS} \cite{liu2018darts}, and deploy the searched networks on three commercial mobile devices; each device is evaluated on both its CPU and GPU, with specifications given in Table~\ref{tab:cpu_gpu_specs}.  Our evaluation follows four steps.  We first confirm that the energy efficiency-driven search space genuinely concentrates architectures with favourable energy-accuracy trade-offs.  We then measure how accurately the kernel-level energy predictors transfer to each new device and how much calibration data they require.  Next, we trace the behaviour of the Pareto-based search procedure described in Algorithm~\ref{algorithm:Pareto_search} and the subsequent model-selection rule in Algorithm~\ref{algorithm:BestModelSelection}.

\begin{table}[t]
\centering
\footnotesize  % Smaller font to fit within single column
\caption{CPU and GPU Specifications of Selected Edge Devices}
\label{tab:cpu_gpu_specs}
\begin{adjustbox}{max width=\columnwidth}
% \resizebox{0.46\textwidth}{!}{
\begin{tabular}{lccc}
\toprule
\textbf{Model} & \textbf{CPU (Cores \& Freq)} &  \textbf{GPU} & \textbf{Vendor} \\
\midrule
Pixel 8 Pro     & 1×X3 + 4×A715 + 4xA510 @ 3GHz & G715s & Google \\
Pixel 7         & 2×X1 + 2×A78 + 4xA55 @ 2.85GHz &  G710   & Google \\
Android Device-3 & 2×A76 + 2×A76 + 4xA55 @ 2.86GHz  &  Mali-G76   & -- \\
\bottomrule
\end{tabular}
% }
\end{adjustbox}
\end{table}

\subsection{Energy Efficiency-driven Search Space Design} We first generate model variants based on three foundational NAS search space. In total, 959,417 valid models are generated for NAS-Bench-201. Similarly, we extend the search spaces of NDS-DARTS and NDS-ENAS from 5,000 to 136,391 and 130,549 valid models, respectively. We demonstrate how the model variant generation effectively increases the likelihood of discovering energy-efficient models. We estimate the energy consumption and accuracy of all models using kernel-level energy predictors trained on Android Device-3 and NASWOT scoring. 

\begin{figure*}[htbp]
  \centering
  \begin{subfigure}[htbp]{0.32\textwidth}
    \includegraphics[width=\linewidth]{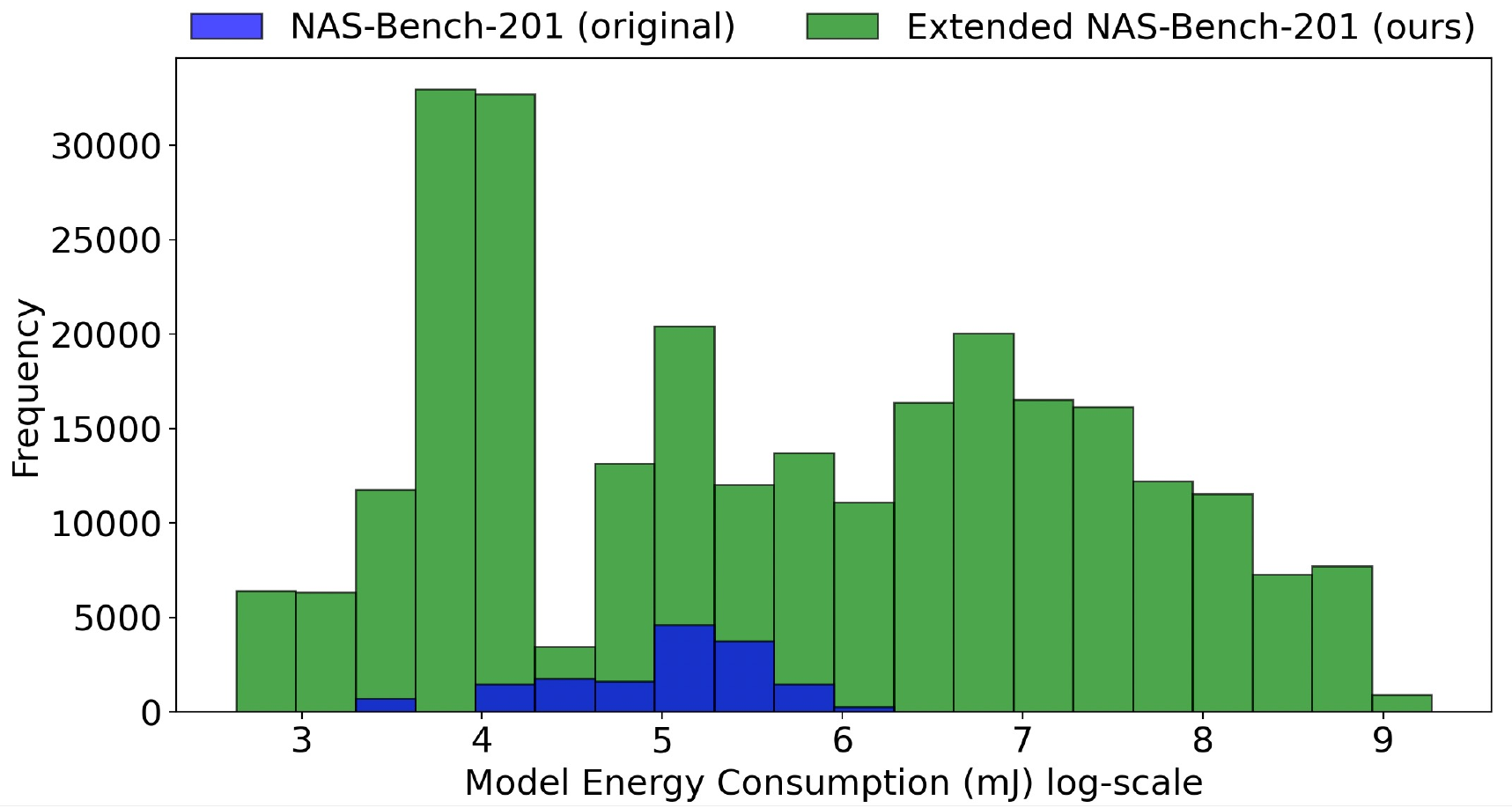}
    \caption{}
    \label{fig:distNASBench201}
  \end{subfigure}
  \hfill
  \begin{subfigure}[htbp]{0.32\textwidth}
    \includegraphics[width=\linewidth]{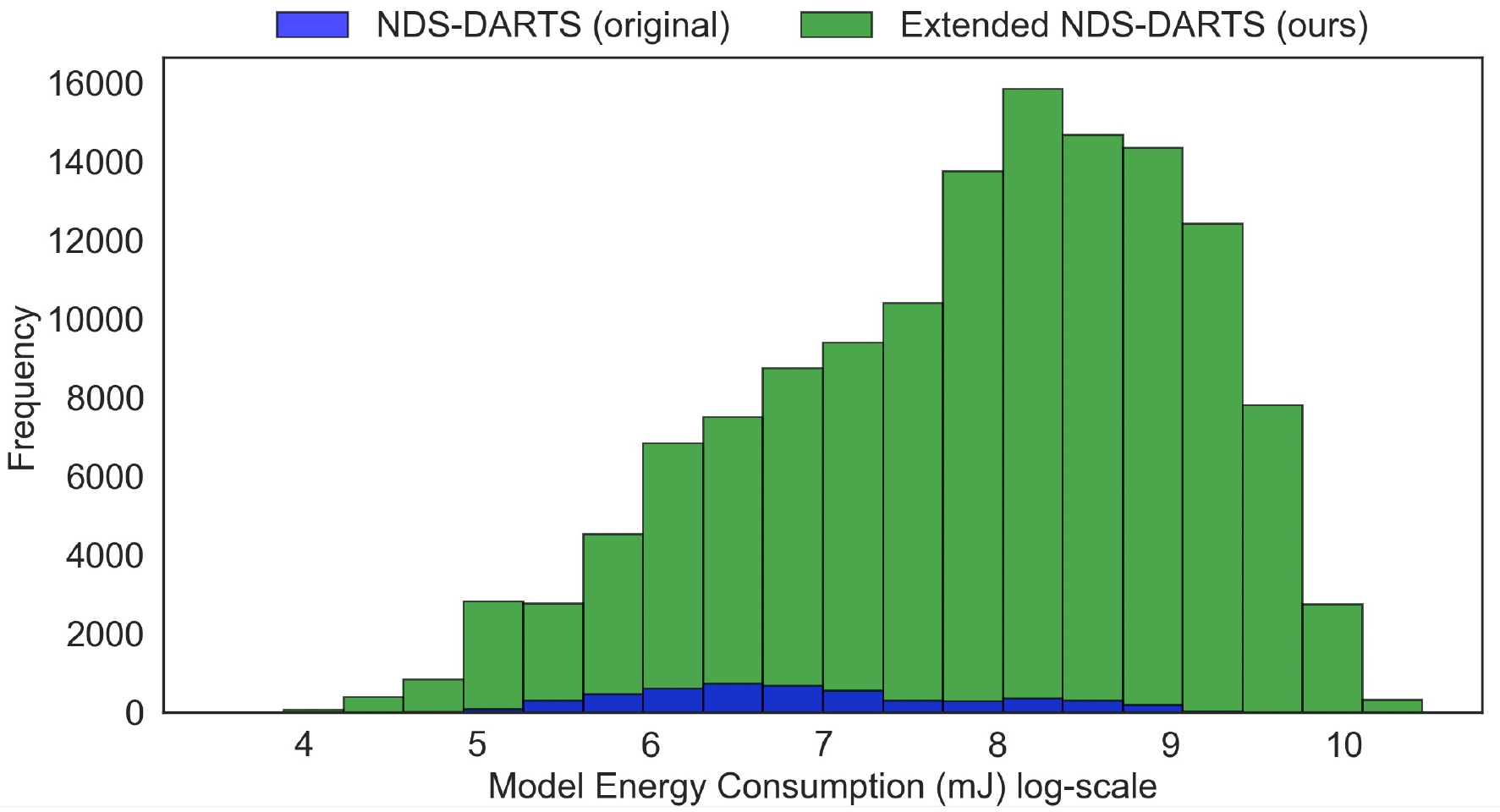}
    \caption{}
    \label{fig:distDARTS}
  \end{subfigure}
  \hfill
  \begin{subfigure}[htbp]{0.32\textwidth}
    \includegraphics[width=\linewidth]{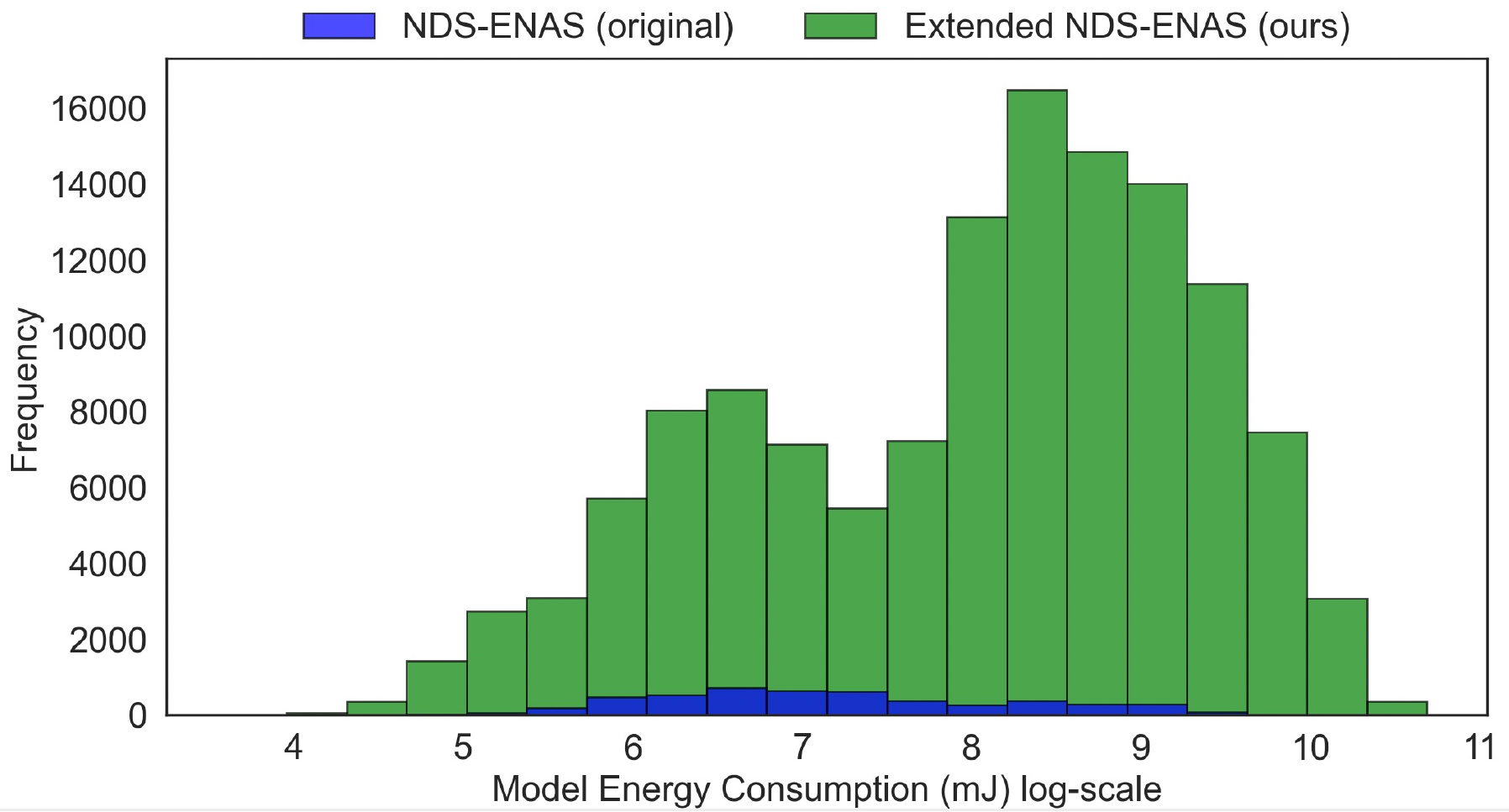}
    \caption{}
    \label{fig:distENAS}
  \end{subfigure}
  \caption{Comparison of model energy distributions evaluated on a mobile \textbf{CPU} before and after search-space extension. (a) NAS-Bench-201. (b) NDS-DARTS. (c) NDS-ENAS.}
  \label{fig:dist_comparison_CPU}
  % \vspace{-0.15in}
\end{figure*}

\begin{figure*}[htbp]
  \centering
  \begin{subfigure}[htbp]{0.32\textwidth}
    \includegraphics[width=\linewidth]{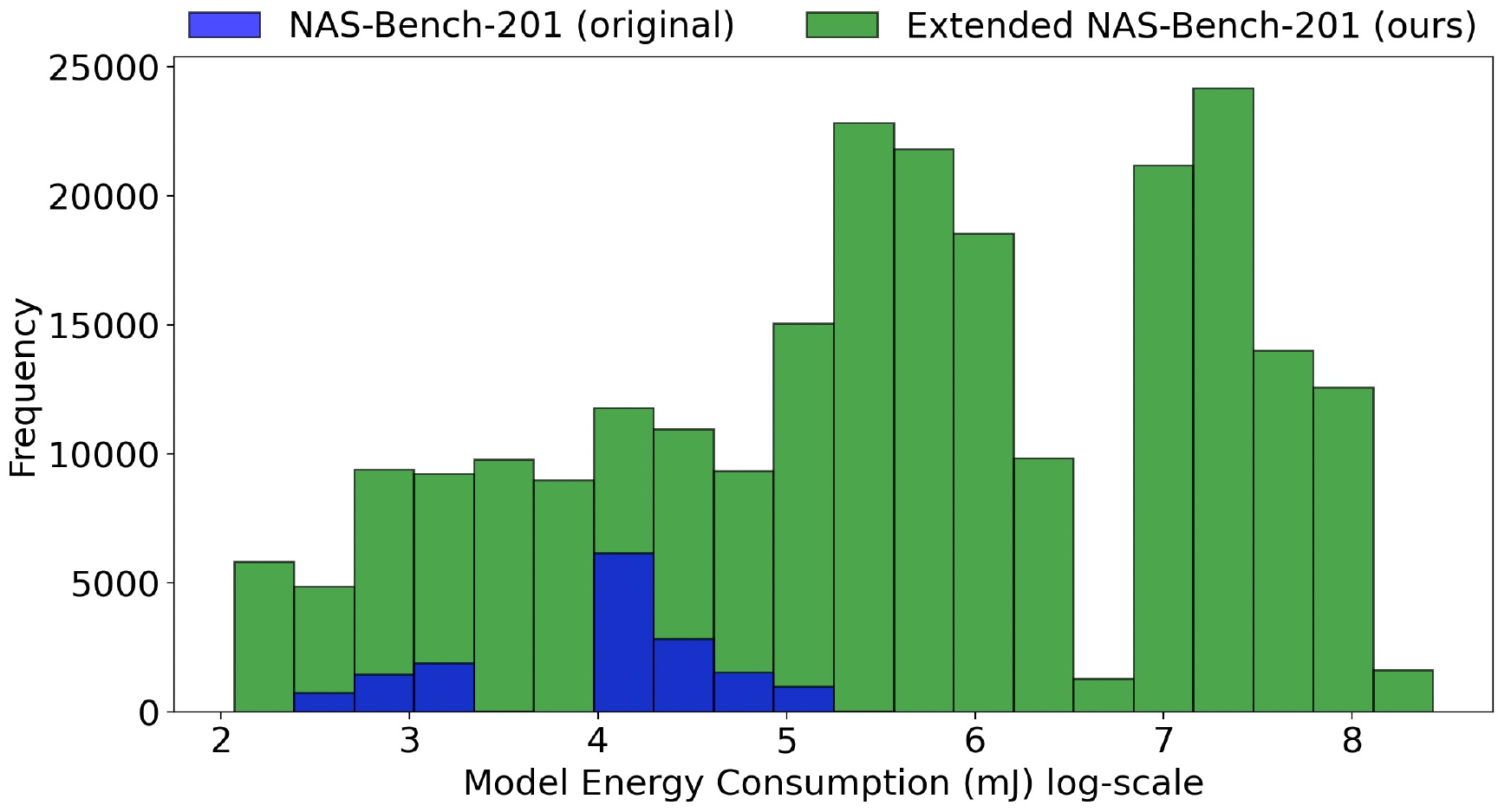}
    \caption{}
    \label{fig:distgpuNASBench201}
  \end{subfigure}
  \hfill
  \begin{subfigure}[htbp]{0.32\textwidth}
    \includegraphics[width=\linewidth]{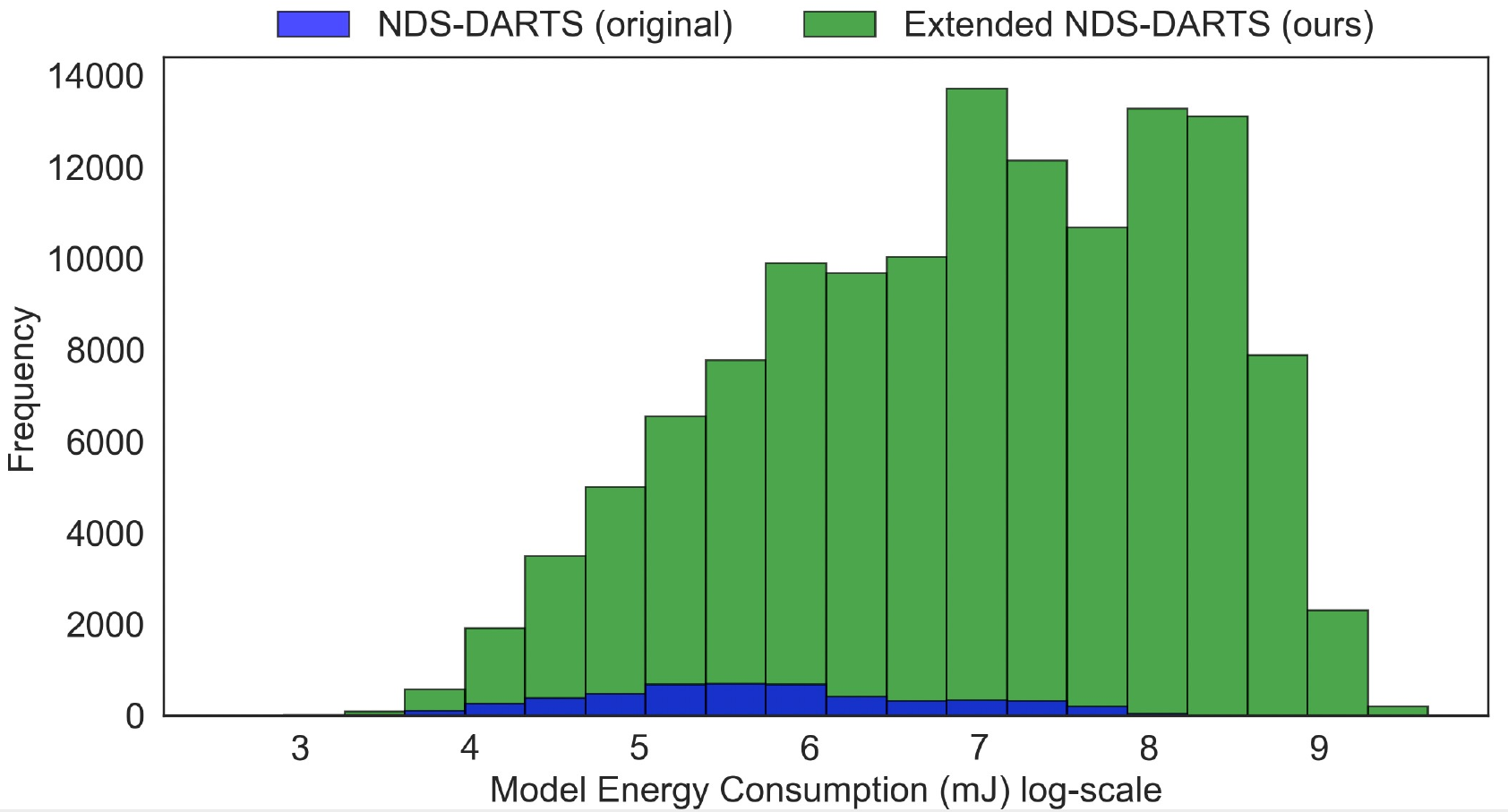}
    \caption{}
    \label{fig:distgpuDARTS}
  \end{subfigure}
  \hfill
  \begin{subfigure}[htbp]{0.32\textwidth}
    \includegraphics[width=\linewidth]{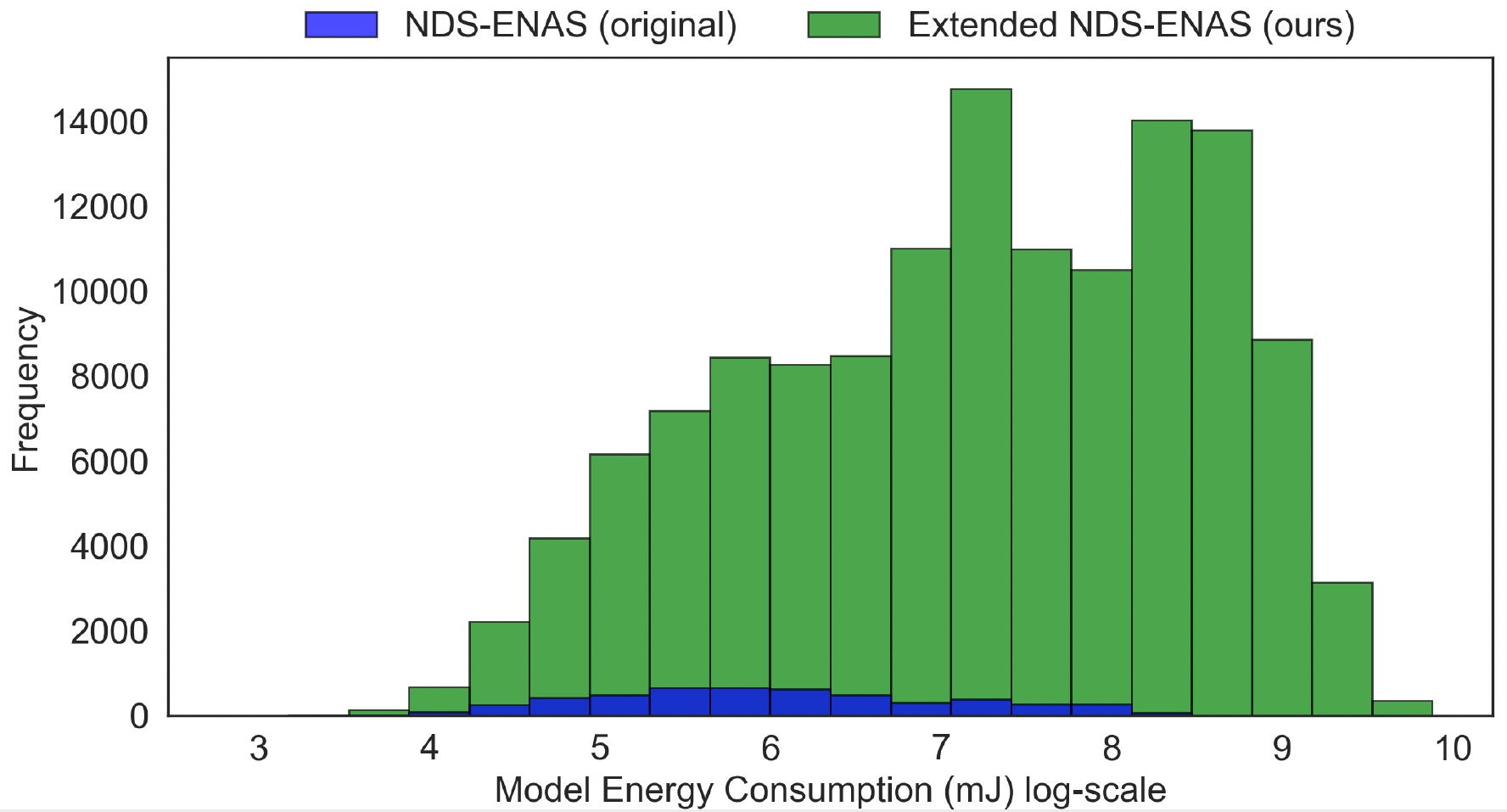}
    \caption{}
    \label{fig:distgpuENAS}
  \end{subfigure}
  \caption{Comparison of model energy distributions evaluated on a mobile \textbf{GPU} before and after search-space extension. (a) NAS-Bench-201. (b) NDS-DARTS. (c) NDS-ENAS.}
  \label{fig:dist_comparison_GPU}
  % \vspace{-0.15in}
\end{figure*}

\begin{figure*}[htbp]
  \centering
  \begin{subfigure}[htbp]{0.32\textwidth}
    \includegraphics[width=\linewidth]{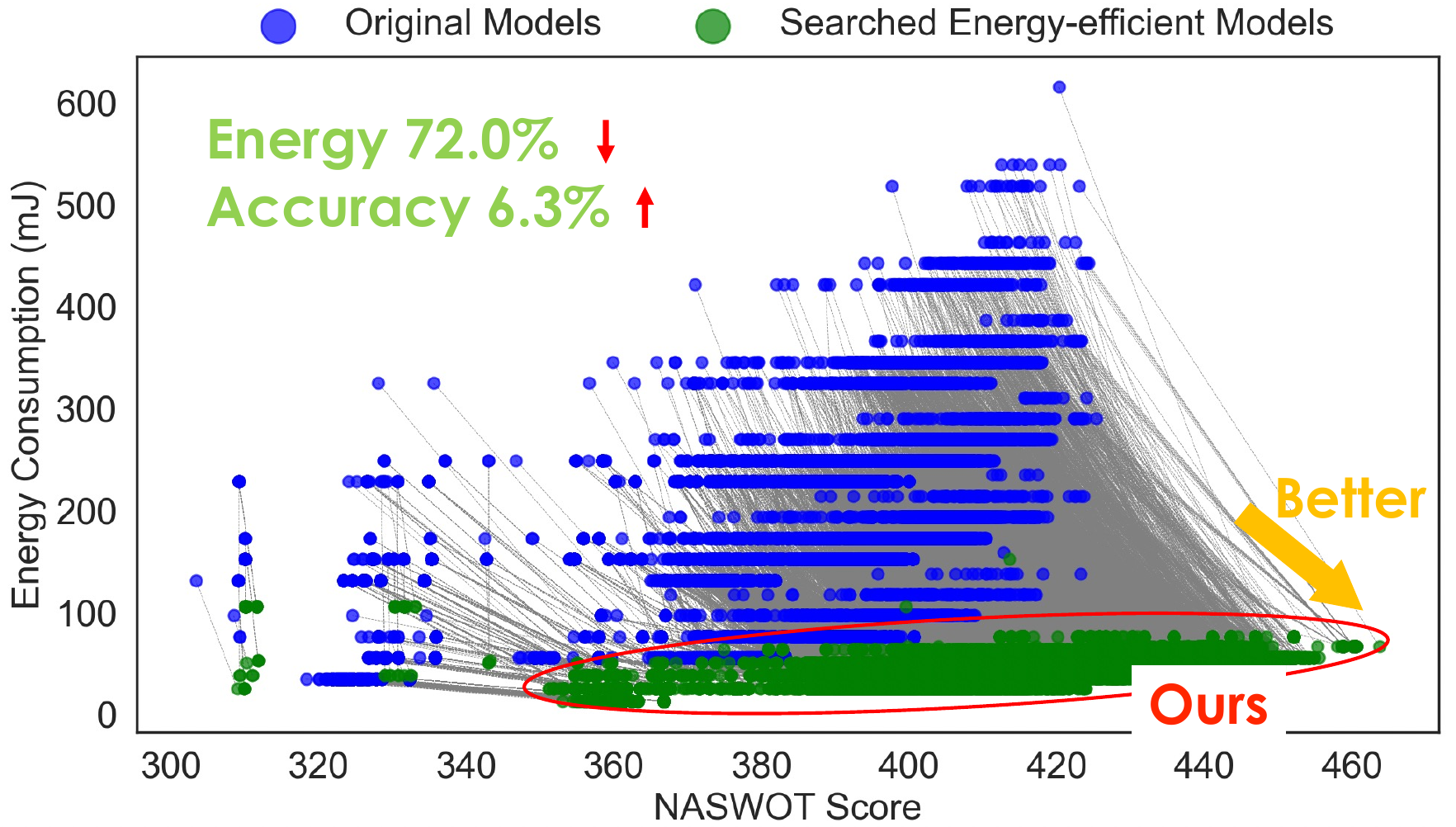}
    \caption{}
    \label{fig:NASBench201}
  \end{subfigure}
  \hfill
  \begin{subfigure}[htbp]{0.32\textwidth}
    \includegraphics[width=\linewidth]{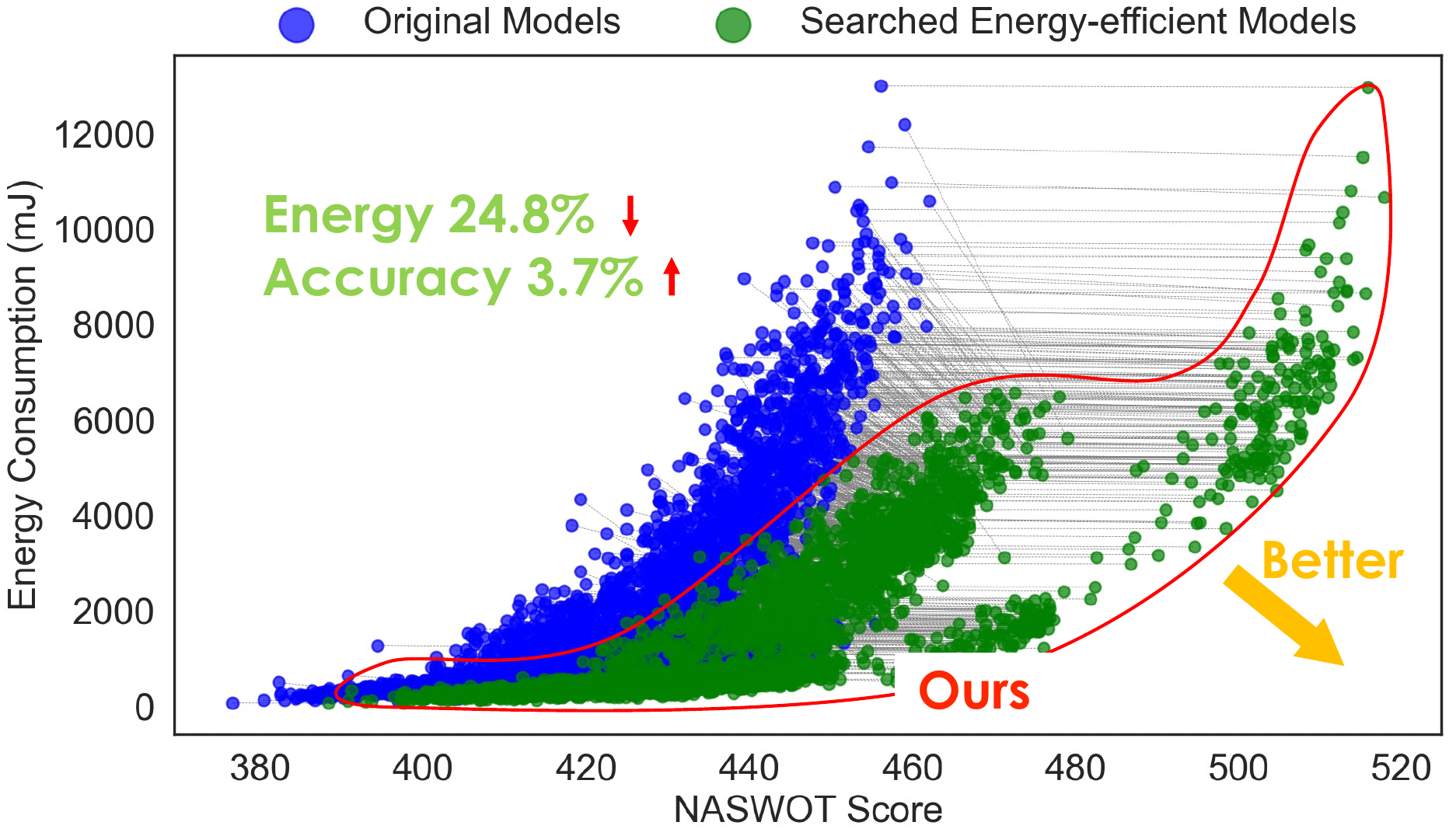}
    \caption{}
    \label{fig:DARTS}
  \end{subfigure}
  \hfill
  \begin{subfigure}[htbp]{0.32\textwidth}
    \includegraphics[width=\linewidth]{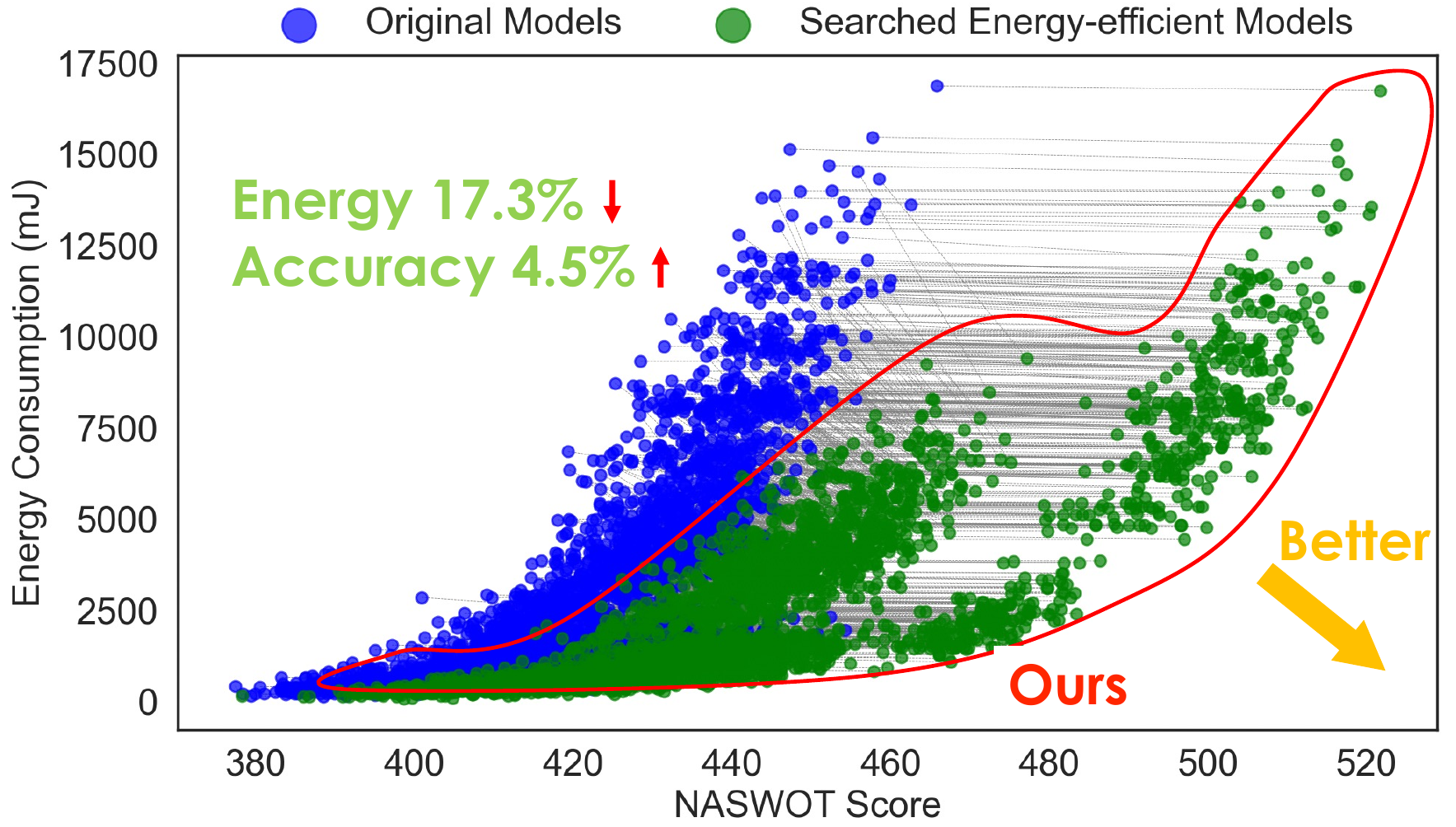}
    \caption{}
    \label{fig:ENAS}
  \end{subfigure}
  \caption{Energy consumption vs.\ NASWOT score on a mobile \textbf{CPU}. Optimized models achieve comparable NASWOT scores with much lower energy. (a) NAS-Bench-201. (b) NDS-DARTS. (c) NDS-ENAS.}
  \label{fig:comparisonCPU}
  % \vspace{-0.15in}
\end{figure*}

\begin{figure*}[htbp]
  \centering
  \begin{subfigure}[htbp]{0.32\textwidth}
    \includegraphics[width=\linewidth]{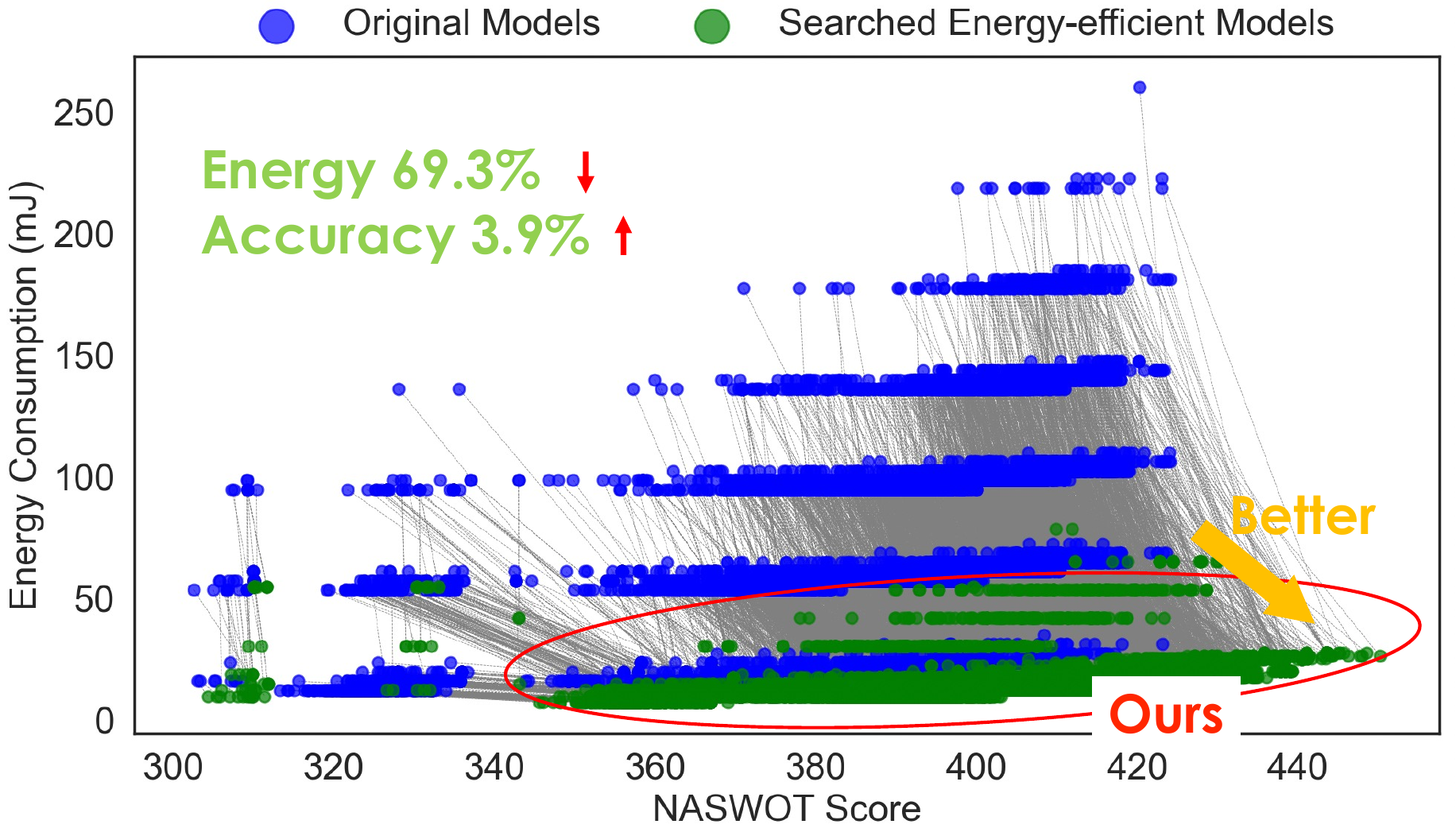}
    \caption{}
    \label{fig:NASBench201gpu}
  \end{subfigure}
  \hfill
  \begin{subfigure}[htbp]{0.32\textwidth}
    \includegraphics[width=\linewidth]{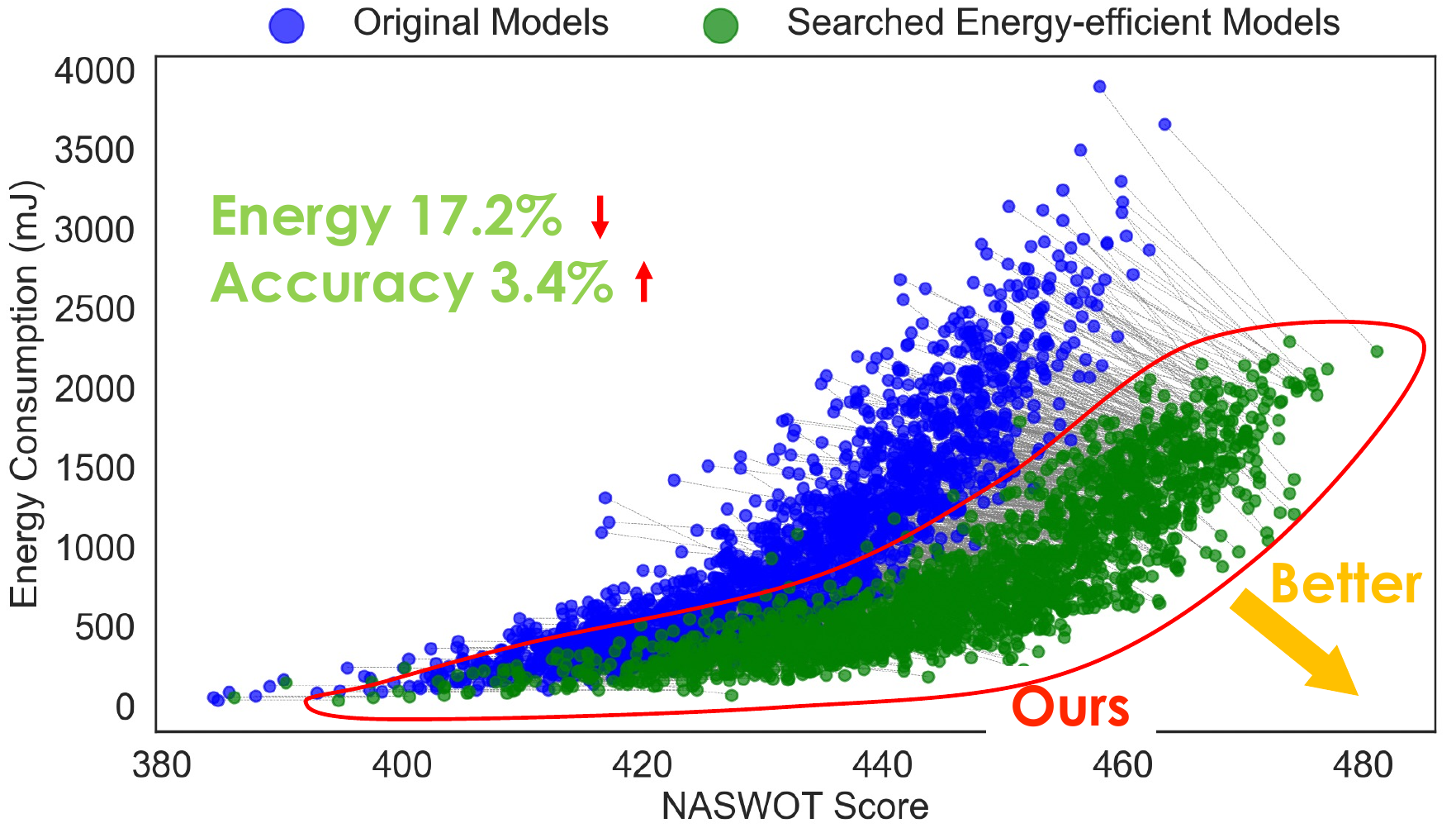}
    \caption{}
    \label{fig:DARTSgpu}
  \end{subfigure}
  \hfill
  \begin{subfigure}[htbp]{0.32\textwidth}
    \includegraphics[width=\linewidth]{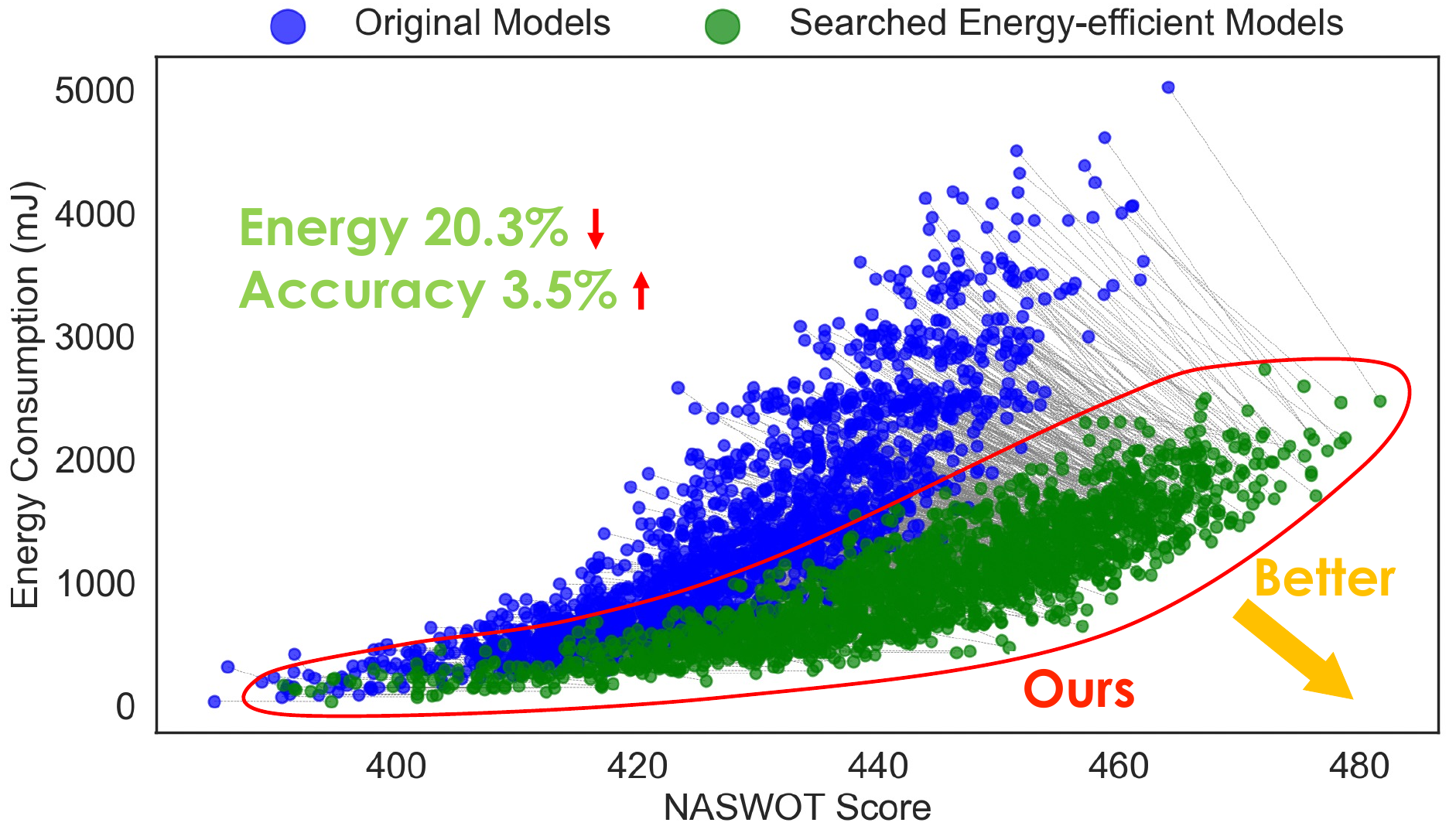}
    \caption{}
    \label{fig:ENASgpu}
  \end{subfigure}
  \caption{Energy consumption vs.\ NASWOT score on a mobile \textbf{GPU}. Optimized models achieve comparable NASWOT scores with much lower energy. (a) NAS-Bench-201. (b) NDS-DARTS. (c) NDS-ENAS.}
  \label{fig:comparisonGPU}
  % \vspace{-0.15in}
\end{figure*}

Fig. \ref{fig:dist_comparison_CPU} and Fig. \ref{fig:dist_comparison_GPU} show the energy distributions of models in the original and in the extended search spaces.The extended space covers a much broader energy range, containing both higher- and lower-consumption architectures than the baseline. This wider spread indicates richer opportunities to locate highly energy-efficient models.
For Fig. \ref{fig:comparisonCPU} and Fig. \ref{fig:comparisonGPU}. We compared the predicted energy consumption and NASWOT Score of all the original architectures from NAS-Bench-201, NDS-DARTS and NDS-ENAS with our optimized models. For each original architecture, we selected an optimized model that maintained similar NASWOT score to its original counterpart but consumed latest energy. Blue points are original models , green points are optimized models. The points near the lower right corner represent the models that achieve a better balance between accuracy and energy cost. Overall, For NAS-Bench-201, the average NASWOT Score of our optimized models increased 6.3\% compared with original models, the average energy consumption decreased 72\%. For NDS-DARTS, the average NASWOT score of our optimized models increased 3.7\% compared with original models, the average energy consumption decreased 24.8\%. For NDS-ENAS, the average NASWOT score of our optimized models increased 4.5\% compared with original models, the average energy consumption decreased 17.3\%. 
In the left figure for NAS-Bench-201, we can see that some architectures have the same predicted energy consumption but different accuracy. This is because these architectures contain the same convolutional operators, and our predictor gives the same energy consumption for them. This suggests a potential improvement for NAS-Bench-201: users can narrow down the search space size to focus more on architectures that offer the best energy and accuracy performance.

From the results, we can see significant improvements in energy consumption with a slight increase in accuracy, especially for NAS-Bench-201. These results demonstrate the great potential of NAS-based model design for exploring energy-efficient models.

\subsection{Transferable Energy Predictors Evaluation}
\label{subsec:eval_predictor}
In this section, we analyze how predictor performance scales with dataset size, introduce strategies to automatically adjust calibration sample sizes, and demonstrate how our predictor adapts efficiently across heterogeneous hardware platforms with minimal effort. 

To assess the generalization capability of our energy predictors across hardware platforms, we first train a base kernel-level predictor on the Google Pixel~7 and Android Device-3. We then adapt this pre-trained model to a new target device, the Pixel~8 Pro, using a small set of additional measurements. This adaptation process is fully automated in \textsc{PlatformX}, enabling scalable energy estimation across diverse edge devices with minimal overhead. We evaluate prediction quality using four metrics: ACC@20, ACC@10, ACC@5, and RMSE.  
ACC@$k$ measures top-$k$ accuracy in a ranking task: given a list of kernel energy values sorted by predicted energy, ACC@$k$ computes the fraction of true top-$k$ lowest-energy kernels correctly identified by the predictor. A higher ACC@$k$ indicates better ranking fidelity for energy-aware selection.  
Root Mean Squared Error (RMSE) quantifies the absolute prediction error in millijoules (mJ), reflecting how closely the predicted energy values align with measured ground truth. A lower RMSE indicates more precise regression performance.

\begin{table*}[t]
\centering
\caption{Accuracy of \texttt{conv+bn+relu} kernel energy prediction when a model trained on Pixel 7 is adapted to Pixel 8 Pro with 30-200 calibration samples.}
\label{tab:transfer_predictor_metrics}
% \footnotesize
% \resizebox{\textwidth}{!}{
\begin{tabular}{llcccccc}
\toprule
Backend & Source Device & Samples & ACC@20 & ACC@10 & ACC@5 & RMSE (mJ) & Time (min) \\ 
\midrule
\multirow{5}{*}{CPU} 
  & Pixel 7 (base)            & 1000 & 86.7\,\% & 47.0\,\% & 16.5\,\% & 221.1 & 40 \\
  & Pixel 8 Pro (adapt) & 30   & 79.8\,\% & 45.4\,\% & 23.6\,\% & 274.2 & 1.2 \\
  & Pixel 8 Pro (adapt) & 50   & 83.0\,\% & 48.2\,\% & 22.5\,\% & 261.1 & 2.0 \\
  & Pixel 8 Pro (adapt) & 100  & \textbf{84.4\,\%} & \textbf{45.4\,\%} & \textbf{22.1\,\%} & 265.2 & 4.0 \\
  & Pixel 8 Pro (adapt) & 200  & 84.0\,\% & 46.6\,\% & 23.6\,\% & 256.1 & 8.0 \\
\midrule
\multirow{5}{*}{GPU} 
  & Pixel 7 (base)            & 1000 & 87.9\,\% & 60.2\,\% & 32.5\,\% & 142.3 & 40 \\
  & Pixel 8 Pro (adapt) & 30   & 68.4\,\% & 38.5\,\% & 18.2\,\% & 193.4 & 1.2 \\
  & Pixel 8 Pro (adapt) & 50   & 73.3\,\% & 39.3\,\% & 19.8\,\% & 150.6 & 2.0 \\
  & Pixel 8 Pro (adapt) & 100  & 73.8\,\% & 38.9\,\% & 21.0\,\% & 148.3 & 4.0 \\
  & Pixel 8 Pro (adapt) & 200  & \textbf{75.5\,\%} & \textbf{41.8\,\%} & \textbf{20.5\,\%} & 141.7 & 8.0 \\
\bottomrule
\end{tabular}
% \vspace{-0.15in}
\end{table*}

\textbf{Training Base Predictors.}  
We first use a VAE-based generative model to synthesize 1{,}000 kernel configurations for the \texttt{conv+bn+relu} operator. 
These configurations are exported as TensorFlow Lite models and deployed to the source devices. 
Kernel-level energy is measured as described in Section~\ref{ssc:measure}. 
Separate MLP predictors are then trained for CPU and GPU backends using these measurements.  
On the source device, the base predictors exhibit strong performance. 
For the CPU, the predictor achieves 86.7\% ACC@20, 47.0\% ACC@10, 16.5\% ACC@5, and an RMSE of 221.1\,mJ. 
For the GPU, it reaches 87.9\% ACC@20, 60.2\% ACC@10, 32.5\% ACC@5, and an RMSE of 142.3\,mJ. 
These results confirm that the trained base predictors effectively capture both ranking fidelity and absolute energy values on their native devices.

\textbf{Cross-Device Adaptation.}  
To evaluate transferability, we adapt the predictors to a Pixel~8 Pro using only 30--200 newly measured samples. 
\textsc{PlatformX} automatically selects representative calibration points and executes them on the target device to collect energy measurements. 
The adaptation completes within four to eight minutes, depending on sample size, and eliminates the need for full retraining.  
Table~\ref{tab:transfer_predictor_metrics} summarizes the results. 
On the CPU backend, direct transfer initially yields reduced accuracy (79.8\% ACC@20, 274.2\,mJ RMSE with 30 samples), but performance improves rapidly as more samples are used. 
With 100 samples, ACC@20 increases to 84.4\%, and RMSE decreases to 265.2\,mJ. 
With 200 samples, the predictor achieves 84.0\% ACC@20 and 256.1\,mJ RMSE.  
A similar trend appears on the GPU backend: with 30 samples, performance starts at 68.4\% ACC@20 and 193.4\,mJ RMSE, improving to 73.8\% ACC@20 and 148.3\,mJ RMSE with 100 samples, and further to 75.5\% ACC@20 and 141.7\,mJ RMSE with 200 samples—comparable to the base model trained with 1{,}000 samples. 
These results confirm that \textsc{PlatformX}'s predictors are highly transferable and sample-efficient, achieving accurate cross-device energy estimation with minimal data. 
This dramatically reduces calibration costs and enables scalable hardware-aware NAS across diverse platforms.

\textbf{Observations and Insights.}  
\textit{More samples do not guarantee higher ranking accuracy.} 
On the CPU path, \textsc{ACC@20} slightly drops from 84.4\% to 84.0\% when the calibration set increases from 100 to 200 samples, even though RMSE improves marginally. 
We attribute this to local overfitting: additional samples refine the regression surface but subtly reshuffle the ordering of low-energy kernels, reducing top-\(k\) recall. 
This suggests that once systematic bias is minimized, blindly increasing calibration data can degrade ranking fidelity. 
A future predictor could therefore employ a stopping criterion or adaptive sampling policy that halts data collection when marginal gains plateau, improving both efficiency and robustness.
\textit{Early diversity matters more than sample count.} 
The GPU predictor benefits more from the first 20 additional samples (30~$\rightarrow$~50) than from the next 150 (50~$\rightarrow$~200) in both \textsc{ACC@20} and RMSE. 
The early samples, selected for operator diversity, contribute richer information, while later randomly drawn samples are often redundant. 
This finding motivates a diversity-aware calibration strategy that prioritizes informative and structurally distinct kernels over raw sample quantity, enabling faster convergence with fewer data points.
\textit{Calibration cost differs by backend.} 
The CPU model requires roughly twice as many samples as the GPU model to achieve comparable \textsc{ACC@20}, yet the total adaptation time differs by only about four minutes. 
Because CPU kernels execute more slowly, each additional sample incurs higher time and energy costs but yields diminishing accuracy improvements. 
This backend asymmetry highlights the need for a resource-aware calibration scheduler that dynamically allocates measurement effort based on backend efficiency, energy cost, and marginal performance gain. 
Such an adaptive scheme would maximize hardware utilization while minimizing unnecessary profiling overhead.

\begin{figure*}[t]
  \centering
  \begin{subfigure}[t]{0.48\linewidth}
    \centering
    \includegraphics[width=\linewidth]{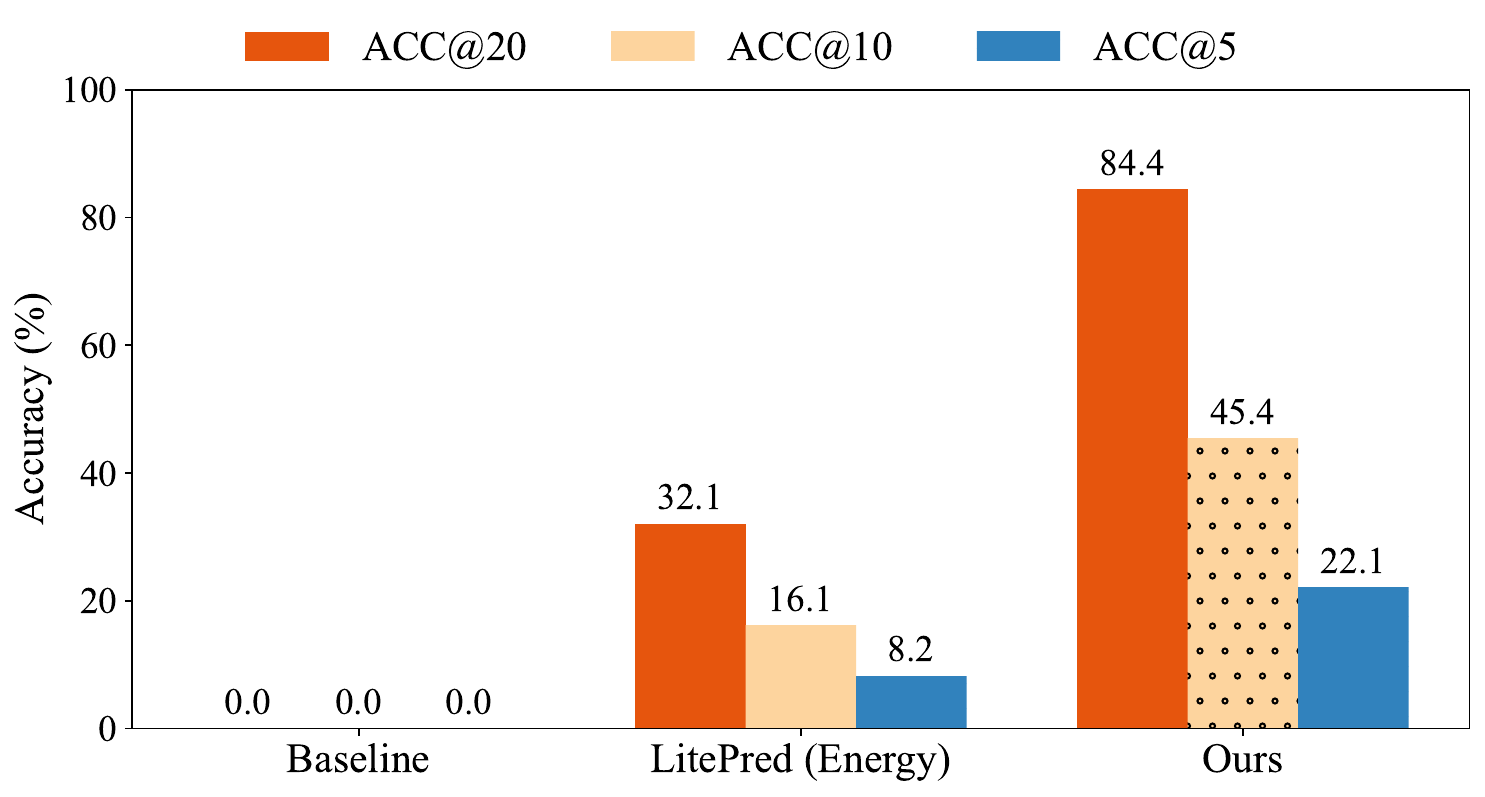}
    \caption{Prediction Performance on Pixel 8 Pro CPU}
    \label{fig:scatter_pixel8_cpu}
  \end{subfigure}%
  \hfill
  \begin{subfigure}[t]{0.48\linewidth}
    \centering
    \includegraphics[width=\linewidth]{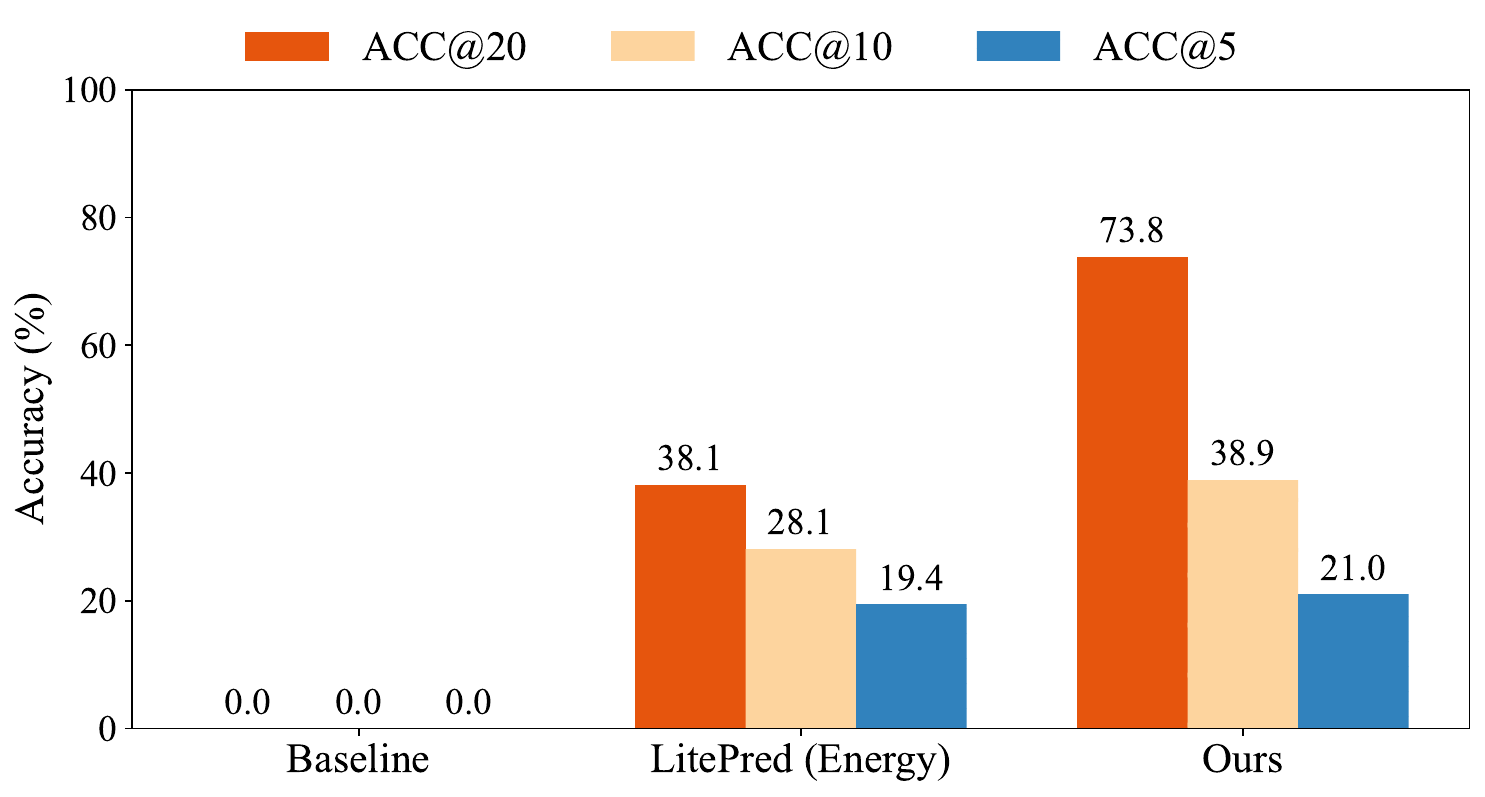}
    \caption{Prediction Performance on Pixel 8 Pro GPU}
    \label{fig:scatter_pixel8_gpu}
  \end{subfigure}
  % \vspace{-0.05in}
    \caption{Energy-prediction accuracy on a Pixel 8 Pro for \texttt{conv+bn+relu} kernels.  
    Results compare our transferable predictor with a device-specific baseline and a modified LitePred, evaluated on both CPU and GPU backends.}
  \label{fig:scatter_pixel8}
  % \vspace{-0.25in}
\end{figure*}

\textbf{Comparison with Prior Predictors.}
To assess the effectiveness of our transferable predictor, we compare it against two baselines: (i) a pre-trained energy predictor from \cite{tu2023unveiling} trained solely on Android Device 3, and (ii) an adaptation of LitePred, modified to regress energy rather than latency.
Fig.~\ref{fig:scatter_pixel8} illustrates the energy prediction error distributions. On the CPU, our adapted predictor achieves 84.4\% ACC@20, 45.4\% ACC@10, and 22.1\% ACC@5, substantially outperforming LitePred (32.1\%, 16.1\%, 8.2\%). On the GPU, the gains are consistent, with our model achieving 73.8\%, 38.9\%, and 21.0\%, compared to LitePred’s 38.1\%, 28.1\%, and 19.4\%. The device-specific baseline predictor performs poorly on the Pixel 8 Pro, with ranking accuracy near zero, which underscores the need for cross-device adaptation. LitePred fares somewhat better, yet its latency-focused design does not transfer well to energy prediction.
These results highlight the practical advantages of our kernel-level predictor with thin, device-specific fine-tuning: it enables fast, accurate energy estimation across hardware platforms, ideal for scalable, energy-constrained hardware-aware NAS workflows.

\subsection{Pareto-based Model Search Evaluation}
\label{subsec:eval_pareto}

We evaluate the model search process on a real mobile device with, and the model search is targeting on the CPU backend. The evaluation covers two part, Pareto-based model sampling and optimal model selection. 

\begin{figure*}[t]
  \centering
  \begin{subfigure}{0.24\textwidth}
    \includegraphics[width=\linewidth]{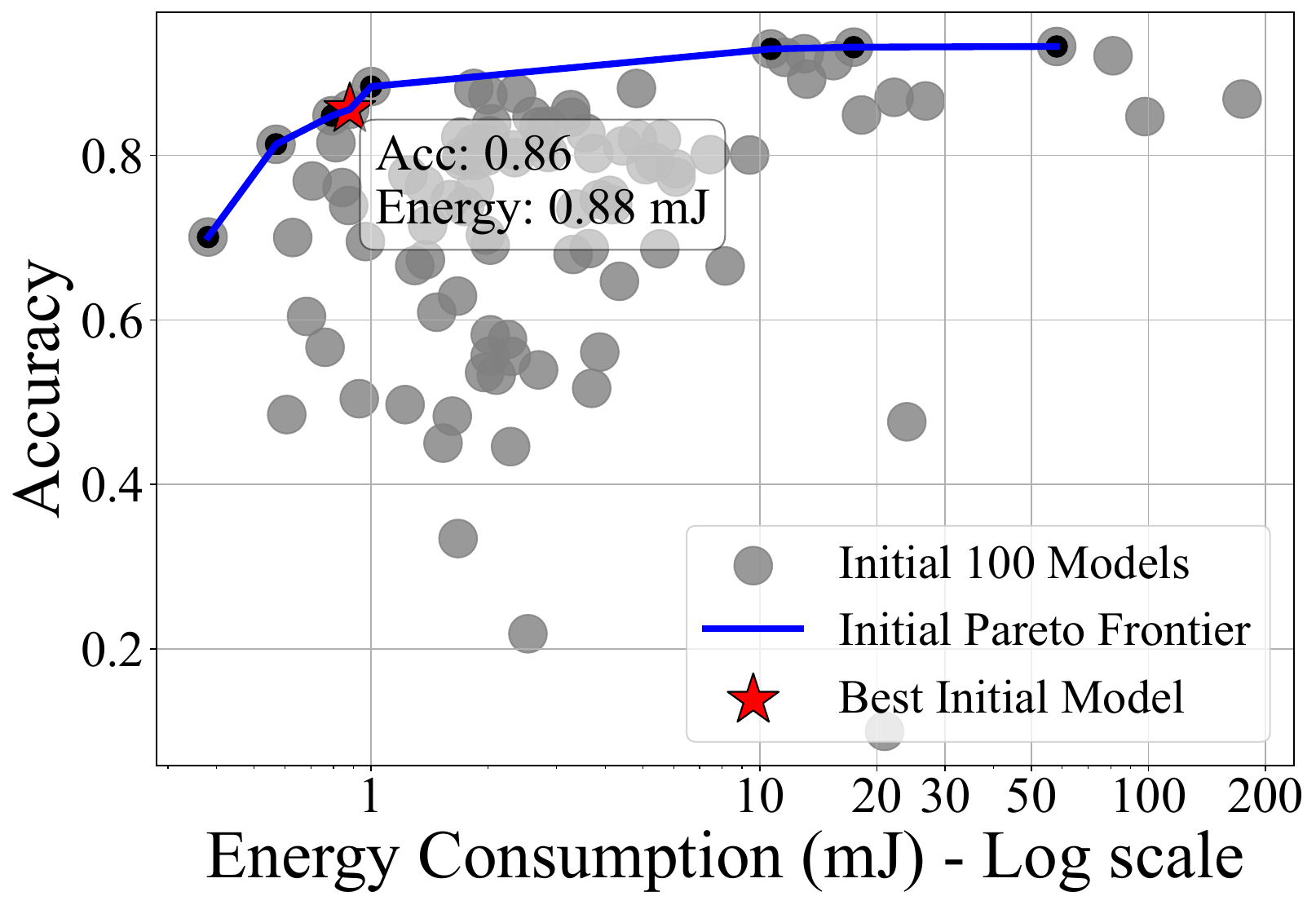}
    \caption{}
    \label{fig:iteration0}
  \end{subfigure}
  \hfill
  \begin{subfigure}{0.24\textwidth}
    \includegraphics[width=\linewidth]{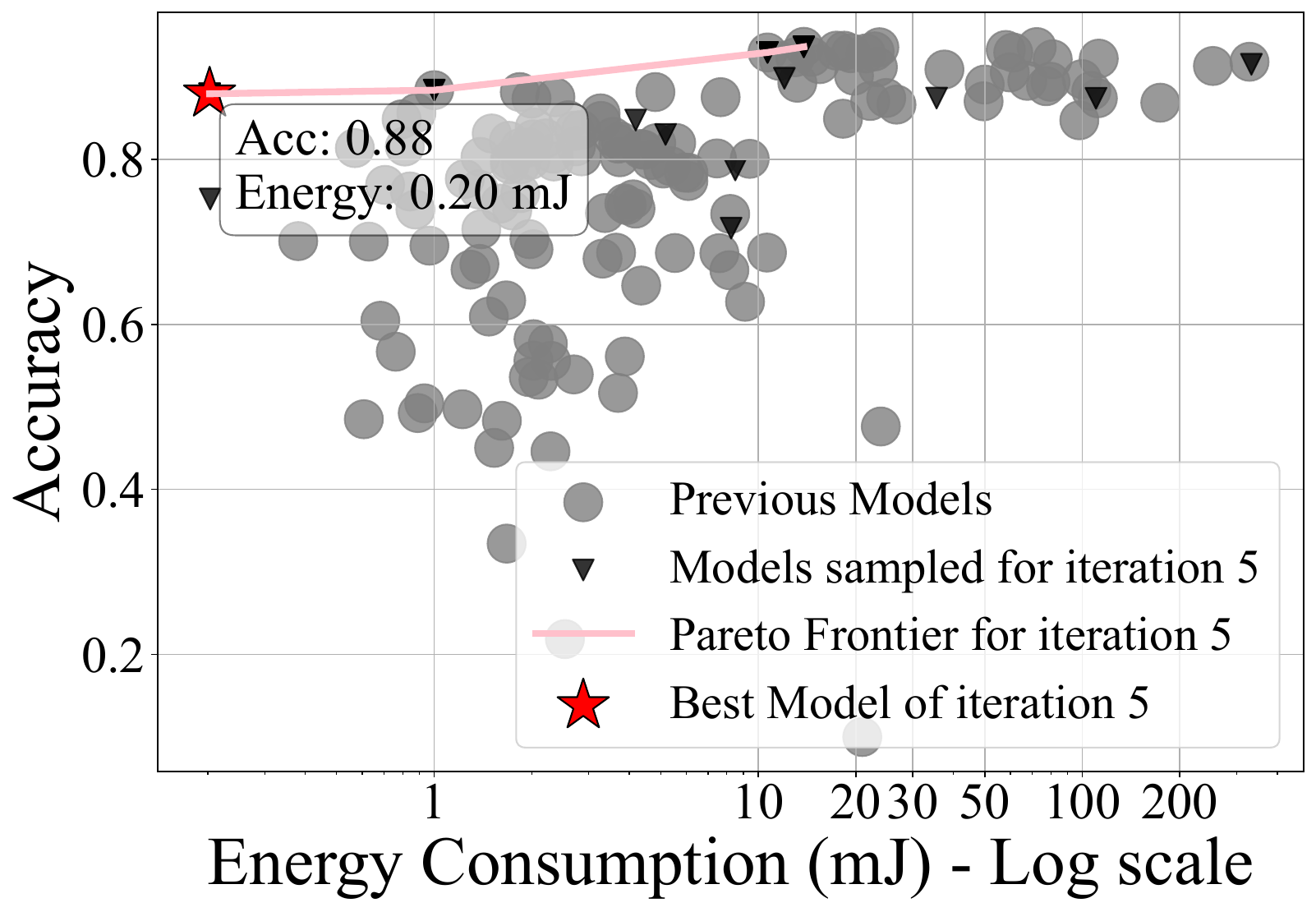}
    \caption{}
    \label{fig:iteration5}
  \end{subfigure}
  \hfill
  \begin{subfigure}{0.24\textwidth}
    \includegraphics[width=\linewidth]{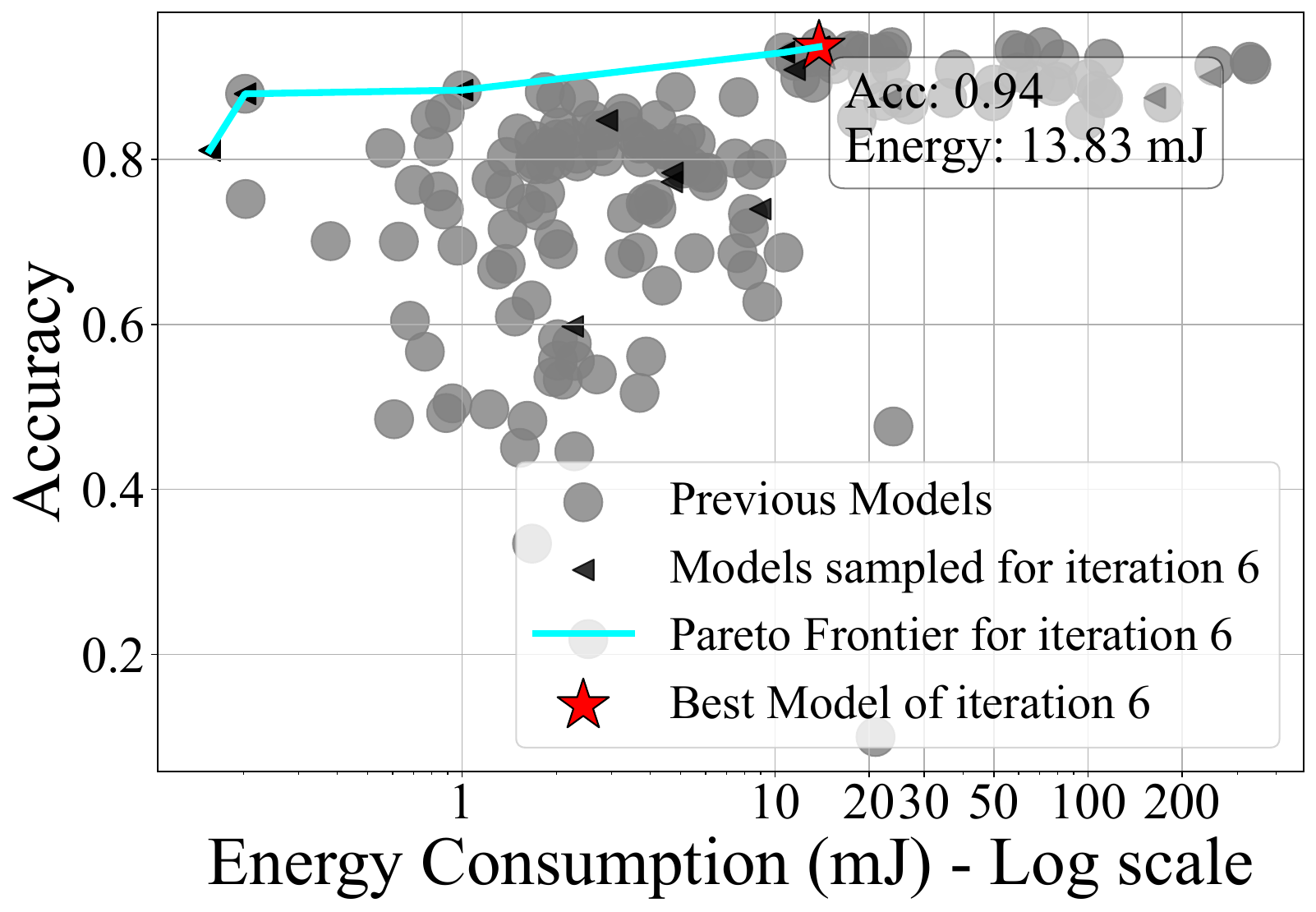}
    \caption{}
    \label{fig:iteration6}
  \end{subfigure}
  \hfill
  \begin{subfigure}{0.24\textwidth}
    \includegraphics[width=\linewidth]{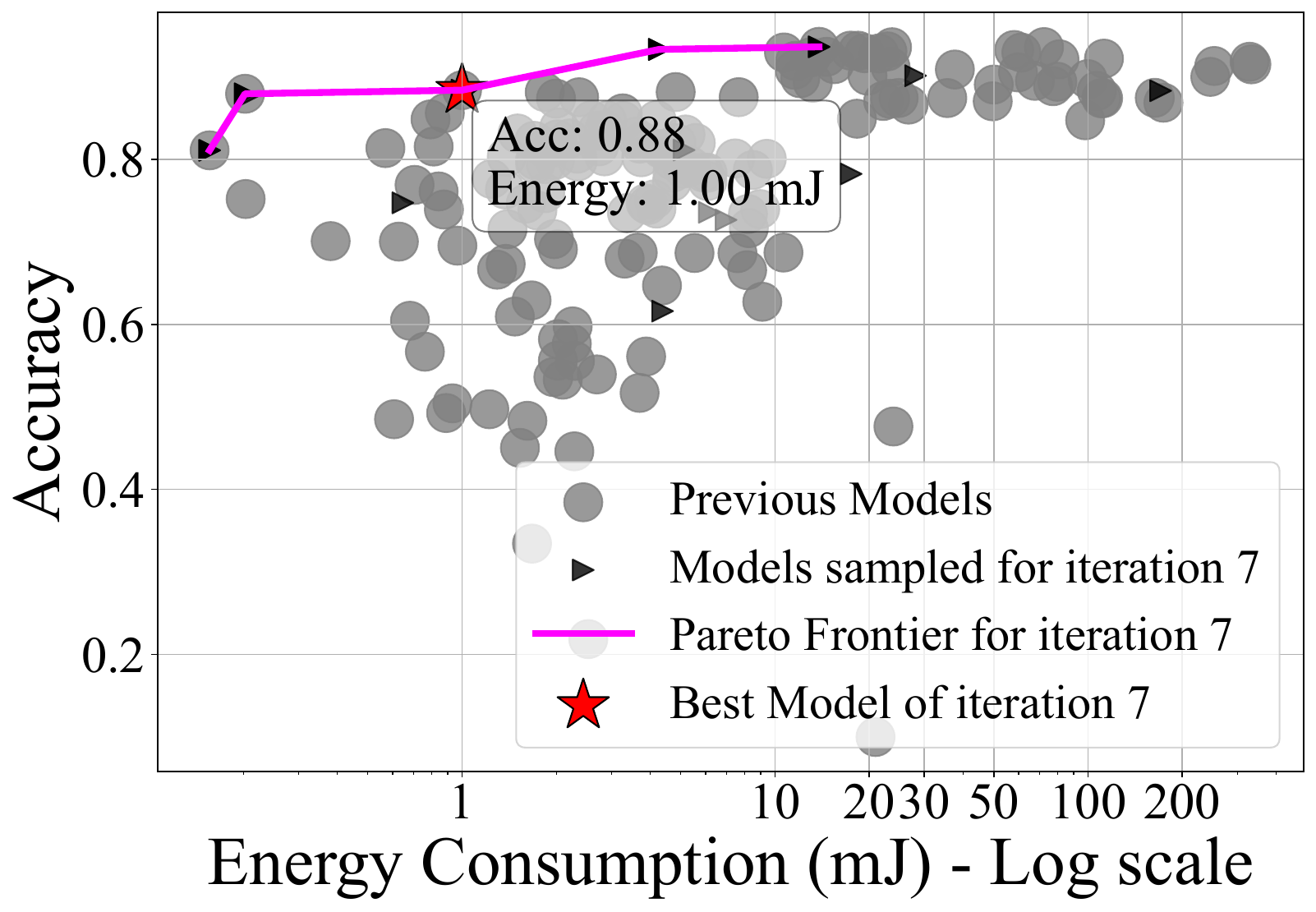}
    \caption{}
    \label{fig:iteration7}
  \end{subfigure}
  % \vspace{-0.1in}
  \caption{Pareto front update and best model selection across multiple search iterations.}
  \label{fig:iterations}
  % \vspace{-0.1in}
\end{figure*}

\textbf{Pareto-based Model Sampling.}
We evaluate our Pareto-based sampling approach on Android Device-3. The evaluation begin with an initial selection of \(m = 100\) models to construct the Pareto front. In each iteration, \(k = 10\) new models are selected for on-device evaluation. The model search weights are configured as \(ws_a = 1\) (accuracy) and \(ws_e = 3\) (energy), encouraging the search to prioritize energy-efficient models. 
Fig.~\ref{fig:iterations} illustrates the search evolution across 7 iterations. Gray points represent the initial models, plotted by energy consumption per 10 million inferences (x-axis) and accuracy (y-axis). The initial Pareto front is shown in blue, and the best model is marked by a red star. After the first iteration, the best model achieves an accuracy of 0.86 and energy consumption of 0.2 mJ.
As shown in Fig.~\ref{fig:iteration5}, 10 new models are evaluated, and the updated Pareto front (yellow line) incorporates real measurements. Although no better model is found in the first iteration, subsequent iterations refine the Pareto front. By the fifth iteration, the best model achieves 0.88 accuracy with only 0.2 mJ energy consumption.
The search direction can be adjusted dynamically by tuning \(ws_i\), allowing the system to favor accuracy or energy efficiency based on user-defined priorities.

\textbf{Optimal Model Selection.}
To assess model selection strategies, we compare Gradient Descent (GD)-based selection (Algorithm~\ref{algorithm:BestModelSelection}) with the traditional Weighted Sum (WS) strategy, using the same pool of measured models. Three scenarios are considered, as shown in Table~\ref{tab:model_selection}:

\begin{itemize}
  \item \textit{Balanced:} \(wd_a = wd_c = 1\)
  \item \textit{Accuracy-prioritized:} \(wd_a = 100, wd_c = 1\)
  \item \textit{Energy-prioritized:} \(wd_a = 1, wd_c = 100\)
\end{itemize}
Under balanced weights, GD identifies a model with 0.88 accuracy and only 0.2~mJ energy consumption within five iterations. When prioritizing accuracy, GD selects a high-performing model with 0.94 accuracy but a much higher energy cost of 13.83~mJ. Conversely, energy-prioritized GD yields a highly efficient model with 0.81 accuracy and just 0.16~mJ per inference.
The WS strategy is less effective across all scenarios. With balanced weights, it selects a model with 0.93 accuracy but significantly higher energy usage (10.68~mJ). In both the accuracy- and energy-prioritized settings, it consistently chooses the same model with 0.88 accuracy and 0.21~mJ energy cost. The best model achieved a validation accuracy of 0.93, with a search overhead of 0.034 kWh per model. In contrast, NASNet-A requires $450$ Nvidia K40 GPUs running continuously for seven days to train 20,000 models, with the best model achieving a slightly lower accuracy of 0.92. However, its search process has a significantly higher search overhead of 0.603 kWh per model. 

Overall, GD-based selection consistently identifies better trade-offs. For each configuration, it finds models closer to the desired optimization goal, outperforming WS in both accuracy and sustainability. Notably, both GD and WS-derived models surpass MobileNet-V2 in accuracy and energy consumption when evaluated on the same edge device.

\begin{table*}[t]
\centering
\caption{Comparison of NAS search results on the CIFAR-10 dataset and the associated energy consumption of the NAS process. }
\label{tab:model_selection}
\footnotesize
\setlength{\tabcolsep}{6pt}   % compact columns
\newcommand{\hdr}[2]{\makecell{#1\\#2}}   % helper for two-line headers
\begin{tabular}{l|l|l|c|c|c|c}
\toprule
Method &
Model Sampling &
Model Selection &
Iterations &
\hdr{Validation Acc.\,$\uparrow$}{(CIFAR-10)} &
\hdr{Inf.\ Energy\,$\downarrow$}{(mJ / inf.)} &
\hdr{Search Overhead\,$\downarrow$}{(kWh / model)} \\
\midrule
\multirow{3}{*}{Gradient Descent (GD)} & \multirow{3}{*}{Energy-prioritized}
  & Balanced              & 5 & 0.88 & 0.21  & \multirow{6}{*}{\textbf{0.034}} \\
 & & Accuracy-prioritized   & 6 & \textbf{0.94} & 13.83 & \\
 & & Energy-prioritized     & 6 & 0.81 & \textbf{0.16} & \\ \cline{1-6}
\multirow{3}{*}{Weighted Sum (WS)}     & \multirow{3}{*}{Energy-prioritized}
  & Balanced              & 5 & \textbf{0.93} & 10.68 & \\
 & & Accuracy-prioritized   & 6 & 0.88 & \textbf{0.21} & \\
 & & Energy-prioritized     & 6 & 0.88 & 0.21  & \\ \midrule
MobileNet-V2 & -- & -- & -- & 0.88 & 7.53 & -- \\
\bottomrule
\end{tabular}
\end{table*}

\section{Related Work}

The growing demand for sustainable AI has intensified efforts to design energy-efficient models, particularly for deployment on resource-constrained edge devices. Prior research has addressed this challenge from two complementary directions: HW-NAS and fine-grained energy profiling.

\textbf{Hardware-Aware Neural Architecture Search.}  
NAS has become a powerful paradigm for automating DNN design. 
Recent work extends NAS to optimize for hardware-specific objectives such as latency, energy, or carbon emissions \cite{gong2019mixed, li2021hw}. KNAS introduces a training-free strategy based on architectural gradients, enabling efficient model evaluation without full training. CE-NAS \cite{zhao2024end} employs a reinforcement learning agent to dynamically adapt GPU usage based on carbon intensity forecasts, balancing sustainability and performance. While these methods incorporate environmental objectives into NAS, they generally rely on simulation-based estimates or coarse-grained profiling. In contrast, \textsc{PlatformX} couples real-device energy profiling with gradient-based Pareto optimization, enabling accurate and energy-aware model selection for practical deployment across heterogeneous edge hardware.

\textbf{Energy Profiling for On-Device Model Execution.}  
Accurate energy measurement is critical for evaluating AI model sustainability on edge devices. 
Tools such as \texttt{pyJoules} \cite{pyJoules} and \texttt{PowerAPI} \cite{PowerAPI} provide real-time energy profiling but remain limited in accuracy, extensibility, or framework compatibility. 
Many existing profilers are tightly integrated with specific software stacks or restricted to certain model layers, which constrains scalability and cross-platform applicability. 
\textsc{PlatformX} addresses these limitations through a modular and extensible profiling pipeline that supports both kernel-level and model-level measurement across diverse operators and hardware. 
Furthermore, it incorporates transferable predictors trained from real measurements and leverages zero-cost proxies for accuracy estimation, enabling rapid, scalable, and energy-aware evaluation without full model retraining.

\section{Discussion and Future Work}\label{sec:discussion}

\textsc{PlatformX} demonstrates that energy-conscious HW-NAS can evolve from a manual, platform-specific process into a fully automated, scalable workflow. Nevertheless, several important research directions remain open for further exploration.

\textbf{Expanding Beyond Convolutional Search Spaces.}  
The current design of \textsc{PlatformX} focuses on convolutional operators and cell-based CNN architectures with energy-relevant design parameters. 
However, emerging edge workloads increasingly rely on transformer-based models for both vision \cite{liu2021swin, mehta2021mobilevit} and language \cite{touvron2023llama, javaheripi2023phi} applications. 
Extending the search space to include attention layers, token mixers, and lightweight transformer modules would enable \textsc{PlatformX} to optimize a broader spectrum of architectures under realistic energy and latency constraints, bridging the gap between CNN-centric and transformer-based model design.

{\sloppy
\textbf{Operator Placement and Hardware Heterogeneity.}  
Modern mobile SoCs integrate heterogeneous compute units—big.LITTLE CPUs, GPUs, NPUs, and DSPs—each exhibiting distinct power–performance trade-offs. 
The current prototype focuses on per-kernel energy prediction within individual backends. 
Future extensions could incorporate backend-aware scheduling and operator placement strategies to more accurately model co-execution and workload migration. 
Integrating real-time energy feedback into compiler frameworks such as TVM \cite{chen2018tvm} could enable joint optimization of model structure \cite{ke2024carboncp} and deployment strategy, advancing toward cross-backend, energy-adaptive execution.
\par}

\textbf{Cross-Device and Scalable Transfer Learning.}  
Future work may explore meta-learning or continual learning strategies to further minimize calibration overhead and enhance generalization to unseen devices and dynamic usage contexts. 
Lightweight federated transfer approaches could allow models trained on different devices to collaboratively update shared predictors without exchanging raw energy traces, preserving privacy while improving robustness and scalability across heterogeneous hardware ecosystems.

\section{Conclusion}
\label{sec:conclusion}

We present \textsc{PlatformX}, the first fully automated framework that enables practical, energy-aware NAS on real edge devices. \textsc{PlatformX} enhances traditional NAS pipelines by enriching the search space with energy-critical configurations, leveraging transferable kernel-level predictors, guiding exploration through gradient-based Pareto optimization, and incorporating high-resolution, hands-free power measurement.
\textsc{PlatformX} reduces per-device calibration time from days to minutes and achieves accurate energy prediction with around 100 adaptation samples when targeting new hardware. It efficiently ranks tens of thousands of architectures and consistently identifies models with superior trade-offs. \textsc{PlatformX} discovers models with up to 0.94 accuracy or as low as 0.16\,mJ per inference, both surpassing MobileNet-V2 in performance and energy efficiency on the same device. 
% These results demonstrate \textsc{PlatformX}’s capability to scale sustainable HW-NAS across diverse edge devices.

\begin{acks}
Research was sponsored by funds from Toyota Motor North America and the Army Research Laboratory and was accomplished under Cooperative Agreement Number W911NF-23-2-0224. The views and conclusions contained in this document are those of the authors and should not be interpreted as representing the official policies, either expressed or implied, of the Army Research Laboratory or the U.S. Government. The U.S. Government is authorized to reproduce and distribute reprints for Government purposes notwithstanding any copyright notation herein.
\end{acks}

\bibliographystyle{unsrt}
% \clearpage
\bibliography{references}

\begin{thebibliography}{10}

\bibitem{wu2019fbnet}
Bichen Wu, Xiaoliang Dai, et~al.
\newblock {FBNet}: Hardware-aware efficient convnet design via differentiable neural architecture search.
\newblock In {\em IEEE Conference on Computer Vision and Pattern Recognition}, 2019.

\bibitem{dong2020nasbench201}
Xuanyi Dong and Yi~Yang.
\newblock {NAS-Bench-201}: Extending the scope of reproducible neural architecture search.
\newblock In {\em International Conference on Learning Representations}, 2020.

\bibitem{li2021hwnasbench}
Hongxu Li et~al.
\newblock {HW-NASBench}: Hardware-aware neural architecture search benchmark.
\newblock In {\em International Conference on Learning Representations}, 2021.

\bibitem{tan2019efficientnet}
Mingxing Tan and Quoc~V. Le.
\newblock Efficientnet: Rethinking model scaling for convolutional neural networks.
\newblock In {\em International Conference on Machine Learning}, 2019.

\bibitem{gu2020brpnas}
Shishuo Gu et~al.
\newblock Brp-nas: Prediction-based nas via bias-reduced pruning.
\newblock In {\em Advances in Neural Information Processing Systems}, 2020.

\bibitem{tu2023unveiling}
Xiaolong Tu, Anik Mallik, Dawei Chen, Kyungtae Han, Onur Altintas, Haoxin Wang, and Jiang Xie.
\newblock Unveiling energy efficiency in deep learning: Measurement, prediction, and scoring across edge devices.
\newblock In {\em 2023 IEEE/ACM Symposium on Edge Computing (SEC)}, pages 80--93. IEEE, 2023.

\bibitem{wang2020energy}
Haoxin Wang, BaekGyu Kim, Jiang Xie, and Zhu Han.
\newblock Energy drain of the object detection processing pipeline for mobile devices: Analysis and implications.
\newblock {\em IEEE Transactions on Green Communications and Networking}, 5(1):41--60, 2020.

\bibitem{yang2022knas}
Haibo Yang et~al.
\newblock Knas: A training-free neural architecture search for on-device speech recognition.
\newblock In {\em IEEE International Conference on Acoustics, Speech and Signal Processing}, 2022.

\bibitem{canziani2019teadnn}
Marco Canziani, Ivan Hubara, and Elias Jarlebring.
\newblock {TEA-DNN}: The quest for time–energy–accuracy co-optimised deep neural networks.
\newblock In {\em ACM/IEEE International Symposium on Low Power Electronics and Design}, 2019.

\bibitem{pham2018efficient}
Hieu Pham et~al.
\newblock Efficient neural architecture search via parameter sharing.
\newblock In {\em International Conference on Machine Learning}, 2018.

\bibitem{zoph2018learning}
Barret Zoph et~al.
\newblock Learning transferable architectures for scalable image recognition.
\newblock In {\em IEEE Conference on Computer Vision and Pattern Recognition}, 2018.

\bibitem{real2019regularized}
Esteban Real, Alok Aggarwal, Yanping Huang, and Quoc~V Le.
\newblock Regularized evolution for image classifier architecture search.
\newblock In {\em Proc. AAAI Conference on Artificial Intelligence}, pages 4780--4789, 2019.

\bibitem{chu2021tunas}
Hang Chu, Zizheng Pan, et~al.
\newblock Tunas: Simplifying the search space of weight-sharing neural architecture search.
\newblock In {\em International Conference on Machine Learning}, 2021.

\bibitem{mellor2021naswot}
Joe Mellor, Jack Turner, Amos Storkey, and Elliot~J. Crowley.
\newblock Neural architecture search without training.
\newblock In {\em International Conference on Machine Learning}, 2021.

\bibitem{tan2019mnasnet}
Mingxing Tan, Bo~Chen, Ruoming Pang, Vijay Vasudevan, Mark Sandler, Andrew Howard, and Quoc~V Le.
\newblock Mnasnet: Platform-aware neural architecture search for mobile.
\newblock In {\em Proceedings of the IEEE/CVF conference on computer vision and pattern recognition}, pages 2820--2828, 2019.

\bibitem{cai2019proxylessnas}
Han Cai, Ligeng Zhu, and Song Han.
\newblock Proxylessnas: Direct neural architecture search on target task and hardware.
\newblock In {\em International Conference on Learning Representations}, 2019.

\bibitem{feng2024litepred}
Chengquan Feng, Li~Lyna Zhang, Yuanchi Liu, Jiahang Xu, Chengruidong Zhang, Zhiyuan Wang, Ting Cao, Mao Yang, and Haisheng Tan.
\newblock $\{$LitePred$\}$: Transferable and scalable latency prediction for $\{$Hardware-Aware$\}$ neural architecture search.
\newblock In {\em 21st USENIX Symposium on Networked Systems Design and Implementation (NSDI 24)}, pages 1463--1477, 2024.

\bibitem{cai2020onceforall}
Han Cai, Chuang Gan, and Song Han.
\newblock Once for all: Train one network and specialize it for efficient deployment.
\newblock In {\em International Conference on Learning Representations}, 2020.

\bibitem{liu2021pilot}
Hanxiao Liu et~al.
\newblock Pi-nas: Predictive iterative neural architecture search.
\newblock In {\em Advances in Neural Information Processing Systems}, 2021.

\bibitem{zhao2024cenas}
Yiyang Zhao, Yunzhuo Liu, Bo~Jiang, and Tian Guo.
\newblock {CE-NAS}: An end-to-end carbon-efficient neural architecture search framework.
\newblock {\em arXiv preprint arXiv:2406.01414}, 2024.

\bibitem{tu2023deepen2023}
Xiaolong Tu, Anik Mallik, Haoxin Wang, and Jiang Xie.
\newblock Deepen2023: Energy datasets for edge artificial intelligence.
\newblock {\em arXiv preprint arXiv:2312.00103}, 2023.

\bibitem{tu2025greenauto}
Xiaolong Tu, Dawei Chen, Kyungtae Han, Onur Altintas, and Haoxin Wang.
\newblock Greenauto: An automated platform for sustainable ai model design on edge devices.
\newblock In {\em Proceedings of the 26th International Workshop on Mobile Computing Systems and Applications}, pages 7--12, 2025.

\bibitem{tu2025aienergy}
Xiaolong Tu, Anik Mallik, Haoxin Wang, and Jiang Xie.
\newblock Aienergy: An energy benchmark for ai-empowered mobile and iot devices.
\newblock 2025.

\bibitem{boubouh2023powerprofiler}
Karim Boubouh and Robert Basmadjian.
\newblock Powerprofiler: Monitoring energy consumption of machine learning algorithms on android mobile devices.
\newblock In {\em ACM International Conference on Future Energy Systems}, 2023.

\bibitem{mallik2023epam}
Anik Mallik, Haoxin Wang, Jiang Xie, Dawei Chen, and Kyungtae Han.
\newblock Epam: A predictive energy model for mobile ai.
\newblock In {\em Icc 2023-ieee international conference on communications}, pages 954--959. IEEE, 2023.

\bibitem{howard2017mobilenets}
Andrew~G Howard, Menglong Zhu, Bo~Chen, Dmitry Kalenichenko, Weijun Wang, Tobias Weyand, Marco Andreetto, and Hartwig Adam.
\newblock Mobilenets: Efficient convolutional neural networks for mobile vision applications.
\newblock {\em arXiv preprint arXiv:1704.04861}, 2017.

\bibitem{cai2018proxylessnas}
Han Cai, Ligeng Zhu, and Song Han.
\newblock Proxylessnas: Direct neural architecture search on target task and hardware.
\newblock {\em arXiv preprint arXiv:1812.00332}, 2018.

\bibitem{yang2018netadapt}
Tien-Ju Yang, Andrew Howard, Bo~Chen, Xiao Zhang, Alec Go, Mark Sandler, Vivienne Sze, and Hartwig Adam.
\newblock Netadapt: Platform-aware neural network adaptation for mobile applications.
\newblock In {\em Proceedings of the European conference on computer vision (ECCV)}, pages 285--300, 2018.

\bibitem{lin2020mcunet}
Ji~Lin, Wei-Ming Chen, Yujun Lin, John Cohn, Chuang Gan, and Song Han.
\newblock Mcunet: Tiny deep learning on iot devices.
\newblock In {\em Advances in Neural Information Processing Systems}, 2020.

\bibitem{cai2019tea}
Lile Cai, Anne-Maelle Barneche, Arthur Herbout, Chuan~Sheng Foo, Jie Lin, Vijay~Ramaseshan Chandrasekhar, and Mohamed M~Sabry Aly.
\newblock Tea-dnn: the quest for time-energy-accuracy co-optimized deep neural networks.
\newblock In {\em 2019 IEEE/ACM International Symposium on Low Power Electronics and Design (ISLPED)}, pages 1--6. IEEE, 2019.

\bibitem{dudziak2020brp}
Lukasz Dudziak, Thomas Chau, Mohamed Abdelfattah, Royson Lee, Hyeji Kim, and Nicholas Lane.
\newblock Brp-nas: Prediction-based nas using gcns.
\newblock {\em Advances in neural information processing systems}, 33:10480--10490, 2020.

\bibitem{wang2020user}
Haoxin Wang and Jiang Xie.
\newblock User preference based energy-aware mobile ar system with edge computing.
\newblock In {\em Proc. IEEE INFOCOM}, pages 1379--1388, 2020.

\bibitem{wang2022leaf+}
Haoxin Wang, BaekGyu Kim, Jiang Xie, and Zhu Han.
\newblock {LEAF}+{AIO}: Edge-assisted energy-aware object detection for mobile augmented reality.
\newblock {\em IEEE Transactions on Mobile Computing}, 22(10):5933--5948, 2022.

\bibitem{wang2017v}
Haoxin Wang, Jiang Xie, and Tao Han.
\newblock V-handoff: A practical energy efficient handoff for 802.11 infrastructure networks.
\newblock In {\em 2017 IEEE International Conference on Communications (ICC)}, pages 1--6. IEEE, 2017.

\bibitem{dong2020bench}
Xuanyi Dong and Yi~Yang.
\newblock {NAS}-{B}ench-201: Extending the scope of reproducible neural architecture search.
\newblock In {\em Proc. ICLR}, 2020.

\bibitem{mellor2021neural}
Joe Mellor, Jack Turner, Amos Storkey, and Elliot~J Crowley.
\newblock Neural architecture search without training.
\newblock In {\em Proc. International conference on machine learning}, pages 7588--7598. PMLR, 2021.

\bibitem{gunantara2018review}
Nyoman Gunantara.
\newblock A review of multi-objective optimization: Methods and its applications.
\newblock {\em Cogent Engineering}, 5(1):1502242, 2018.

\bibitem{kang2024survey}
Shida Kang, Kaiwen Li, and Rui Wang.
\newblock A survey on pareto front learning for multi-objective optimization.
\newblock {\em Journal of Membrane Computing}, pages 1--7, 2024.

\bibitem{krizhevsky2009learning}
Alex Krizhevsky, Geoffrey Hinton, et~al.
\newblock Learning multiple layers of features from tiny images.
\newblock 2009.

\bibitem{tflitebenchmarktool}
{TFLite Benchmark}.
\newblock \url{https://ai.google.dev/edge/litert/models/measurement}.
\newblock Accessed on Oct. 2024.

\bibitem{Monsoon}
{Monsoon}.
\newblock \url{https://www.msoon.com/high-voltage-power-monitor}.
\newblock Accessed on Oct. 2024.

\bibitem{liu2018darts}
Hanxiao Liu, Karen Simonyan, and Yiming Yang.
\newblock Darts: Differentiable architecture search.
\newblock {\em arXiv preprint arXiv:1806.09055}, 2018.

\bibitem{gong2019mixed}
Chengyue Gong, Zixuan Jiang, Dilin Wang, Yibo Lin, Qiang Liu, and David~Z Pan.
\newblock Mixed precision neural architecture search for energy efficient deep learning.
\newblock In {\em 2019 IEEE/ACM International Conference on Computer-Aided Design (ICCAD)}, pages 1--7. IEEE, 2019.

\bibitem{li2021hw}
Hongxu Li et~al.
\newblock Hw-nasbench: Hardware-aware neural architecture search benchmark.
\newblock In {\em ICLR}, 2021.

\bibitem{zhao2024end}
Yiyang Zhao, Yunzhuo Liu, Bo~Jiang, and Tian Guo.
\newblock Ce-nas: An end-to-end carbon-efficient neural architecture search framework.
\newblock {\em arXiv preprint arXiv:2406.01414}, 2024.

\bibitem{pyJoules}
{pyJoules}.
\newblock \url{https://github.com/powerapi-ng/pyJoules/}.
\newblock Accessed on Oct. 2024.

\bibitem{PowerAPI}
{PowerAPI}.
\newblock \url{https://powerapi.org/}.
\newblock Accessed on Oct. 2024.

\bibitem{liu2021swin}
Ze~Liu, Yutong Lin, Yue Cao, Han Hu, Yixuan Wei, Zheng Zhang, Stephen Lin, and Baining Guo.
\newblock Swin transformer: Hierarchical vision transformer using shifted windows.
\newblock In {\em Proceedings of the IEEE/CVF international conference on computer vision}, pages 10012--10022, 2021.

\bibitem{mehta2021mobilevit}
Sachin Mehta and Mohammad Rastegari.
\newblock Mobilevit: Light-weight, general-purpose, and mobile-friendly vision transformer. arxiv 2021.
\newblock {\em arXiv preprint arXiv:2110.02178}, 2021.

\bibitem{touvron2023llama}
Hugo Touvron, Thibaut Lavril, Gautier Izacard, Xavier Martinet, Marie-Anne Lachaux, Timoth{\'e}e Lacroix, Baptiste Rozi{\`e}re, Naman Goyal, Eric Hambro, Faisal Azhar, et~al.
\newblock Llama: Open and efficient foundation language models.
\newblock {\em arXiv preprint arXiv:2302.13971}, 2023.

\bibitem{javaheripi2023phi}
Mojan Javaheripi, S{\'e}bastien Bubeck, Marah Abdin, Jyoti Aneja, Sebastien Bubeck, Caio C{\'e}sar~Teodoro Mendes, Weizhu Chen, Allie Del~Giorno, Ronen Eldan, Sivakanth Gopi, et~al.
\newblock Phi-2: The surprising power of small language models.
\newblock {\em Microsoft Research Blog}, 1(3):3, 2023.

\bibitem{chen2018tvm}
Tianqi Chen, Thierry Moreau, Ziheng Jiang, Lianmin Zheng, Eddie Yan, Haichen Shen, Meghan Cowan, Leyuan Wang, Yuwei Hu, Luis Ceze, et~al.
\newblock $\{$TVM$\}$: An automated $\{$End-to-End$\}$ optimizing compiler for deep learning.
\newblock In {\em 13th USENIX Symposium on Operating Systems Design and Implementation (OSDI 18)}, pages 578--594, 2018.

\bibitem{ke2024carboncp}
Hongyu Ke, Wanxin Jin, and Haoxin Wang.
\newblock Carboncp: Carbon-aware dnn partitioning with conformal prediction for sustainable edge intelligence.
\newblock {\em arXiv preprint arXiv:2404.16970}, 2024.

\end{thebibliography}

\end{document}